\pdfoutput=1
\PassOptionsToPackage{table,HTML}{xcolor}
\documentclass[11pt]{article}

\usepackage{EMNLP2023}

\usepackage{times}
\usepackage{amsmath} 
\usepackage{ulem}
\usepackage{arydshln} 
\usepackage{latexsym}
\usepackage{multirow}
\usepackage{array}
\usepackage{booktabs}
\usepackage{tabularx} 
\usepackage{graphicx}
\usepackage{ulem}
\usepackage{subcaption}
\usepackage{float}
\usepackage{wrapfig}
\usepackage{afterpage}
\usepackage{float}

\usepackage[T1]{fontenc}

\usepackage[utf8]{inputenc}

\usepackage{microtype}

\usepackage{inconsolata}
\usepackage{tikz}
\usepackage{arydshln} 
\usetikzlibrary{calc}

\title{Do LVLMs Know What They Know? 

A Systematic Study of Knowledge Boundary Perception in LVLMs}

\author{
Zhikai Ding\textsuperscript{\rm 1} \thanks{~~Work done during an internship at ICT,CAS.} \quad
Shiyu Ni\textsuperscript{\rm 1,2} \quad
Keping Bi\textsuperscript{\rm 1,2} \thanks{~~Corresponding author} \\
\textsuperscript{\rm 1} State Key Laboratory of AI Safety, Institute of Computing Technology, Chinese Academy of Sciences \\
\textsuperscript{\rm 2} University of Chinese Academy of Sciences \\
\texttt{dingzhikai158@gmail.com} \\
\texttt{\{nishiyu23z,bikeping\}@ict.ac.cn}
}

\definecolor{singlestep}{RGB}{214, 246, 249}
\definecolor{doublestep}{RGB}{252, 239, 189}
\definecolor{consistency}{RGB}{241, 249, 154}
\definecolor{PPL}{RGB}{253, 161, 229}
\newlength{\boxwidth} 
\newlength{\boxheight} 
\newlength{\boxwidthh}
\begin{document}
\maketitle
\begin{abstract}

Large vision-language models (LVLMs) demonstrate strong visual question answering (VQA) capabilities but are shown to hallucinate. A reliable model should perceive its knowledge boundaries—knowing what it knows and what it does not. This paper investigates LVLMs’ perception of their knowledge boundaries by evaluating three types of confidence signals: probabilistic confidence, answer consistency-based confidence, and verbalized confidence. Experiments on three LVLMs across three VQA datasets show that, although LVLMs possess a reasonable perception level, there is substantial room for improvement. Among the three confidences, probabilistic and consistency-based signals are more reliable indicators, while verbalized confidence often leads to overconfidence. 
To enhance LVLMs’ perception, we adapt several established confidence calibration methods from Large Language Models (LLMs) and propose three effective methods. 
Additionally, we compare LVLMs with their LLM counterparts, finding that jointly processing visual and textual inputs decreases question-answering performance, but reduces confidence, resulting in an improved perception level compared to LLMs. \looseness=-1

\end{abstract}

\section{Introduction}
Large vision-language models (LVLMs) are capable of processing both textual and visual information simultaneously, demonstrating strong performance on visual question-answering (VQA) task \citep{bai2025qwen25vltechnicalreport,wu2024deepseekvl2mixtureofexpertsvisionlanguagemodels,openai2024gpt4technicalreport}. However, when faced with questions beyond their knowledge boundaries, LVLMs often hallucinate---generating seemingly plausible but factually incorrect responses \citep{liu2024surveyhallucinationlargevisionlanguage,bai2025hallucinationmultimodallargelanguage}. This is unacceptable in safety-critical domains such as healthcare.
Knowing when an LVLM can answer correctly not only helps us determine when to trust the model but also enables adaptive retrieval-augmented generation (RAG), triggering RAG only when the model does not know the answer, which improves both the efficiency and effectiveness of RAG \citep{ni2024llmsneedretrievalaugmentation}.

A trustworthy model should have a clear perception of its knowledge boundaries—knowing what it knows and what it does not. While this ability has been extensively studied in large language models (LLMs) \citep{xiong2024llmsexpressuncertaintyempirical,tian2023justaskcalibrationstrategies,moskvoretskii2025adaptiveretrievalselfknowledgebringing}, it remains underexplored in LVLMs. 
A model’s perception level is assessed by the alignment between its confidence and actual performance, with correctness of the answer serving as a proxy for performance. Therefore, the emphasis is on whether LVLMs can provide confidence that matches their performance.
We focus on binary confidence because it directly helps us decide whether to trust the model.


In this work, we explore this question by examining three representative types of confidence signals that are widely used in LLMs: 
\textbf{1) Probabilistic confidence}~\citep{desai2020calibrationpretrainedtransformers, guo2017calibration}. The confidence is measured by the generation probability of tokens in the output. 
\textbf{2) Answer consistency-based confidence}~\citep{zhang2024sac3reliablehallucinationdetection, manakul2023selfcheckgpt}. Some studies argue that token-level probabilities poorly reflect a model’s semantic confidence and are not suitable for black-box models. Instead, they suggest using semantic consistency across multiple responses as a confidence indicator. 
\textbf{3) Verbalized confidence}~\citep{lin2022teaching, yang2024alignmenthonesty}. The natural language confidence expressed by the model, offering an intuitive and model-agnostic signal without requiring repeated sampling.

We conduct experiments using three representative models—-Qwen2.5-VL \citep{bai2025qwen25vltechnicalreport}, DeepSeek-VL2 \citep{wu2024deepseekvl2mixtureofexpertsvisionlanguagemodels}, and LLaVA-v1.5 \citep{liu2024improvedbaselinesvisualinstruction}—on three datasets: Dyn-VQA \citep{li2025benchmarkingmultimodalretrievalaugmented}, MMMU Pro \citep{yue2024mmmuprorobustmultidisciplinemultimodal}, and Visual7W \citep{Zhu_2016_CVPR}.
The results show that LVLMs can perceive their knowledge boundaries to some extent, but there is still considerable room for improvement. Among the three types of confidence, probabilistic and answer consistency-based confidences are more aligned with LVLMs' performance but rely on in-domain data for binarization, while verbalized confidence has weaker alignment and tends to be overconfident.

To enhance LVLMs' perception capabilities, we adopt several representative confidence calibration methods originally designed for LLMs. The results show that methods that engage the model's reasoning abilities can enhance both answer accuracy and verbalized perception level, whereas existing consistency-based methods have limited effectiveness and do not generalize well to LVLMs. 
We also propose three new approaches tailored for LVLMs: Img-CoT, Prob-Thr, and Cross Model, making further explorations into measuring their confidence. \looseness=-1

Compared to LLMs, LVLMs need to process an additional visual modality and integrate information across different modalities. This raises the question: How does the perception ability of LVLMs differ from that of LLMs? 
To investigate this, we compare the LVLMs with their corresponding LLMs on the Dyn-VQA dataset \citep{li2025benchmarkingmultimodalretrievalaugmented,tian2023justaskcalibrationstrategies}. This dataset provides parallel visual-textual and pure textual queries, ensuring fair comparison between LLMs and LVLMs. We focus on verbalized confidence because it can reflect the model's language capabilities.
Experimental results show that: 1) LVLMs exhibit lower VQA performance but higher perception accuracy compared to their LLM counterparts. 2) Certain prompting methods are ineffective for LVLMs, showing that LVLMs have weaker instruction-following capabilities. 

We hypothesize these phenomena may be caused by the following two reasons: 
1) Compared to processing single-modality information, handling visual inputs and integrating multiple modalities is more challenging for LVLMs. This results in lower VQA performance but also reduces the models’ confidence, mitigating overconfidence and yielding a more accurate perception of their abilities.
2) Training LLMs without sufficient capacity to accommodate additional visual information can erode their language abilities, thereby weakening their instruction-following skills.
Controlled experiments across different model scales and input modalities support these hypotheses.


\section{Related Work}

\textbf{LLM Knowledge Boundary Perception.} Prior research has primarily focused on knowledge boundary perception in LLMs, with various methodologies proposed to elicit confidence: verbalized confidence, where models directly articulate their confidence \citep{yang2024alignmenthonesty,yin2023largelanguagemodelsknow,zhang2023pacelmpromptingaugmentationcalibrated}; consistency based confidence that derive confidence from answer consistency across multiple samples \citep{manakul2023selfcheckgptzeroresourceblackboxhallucination,agrawal2024languagemodelsknowtheyre}; probabilistic confidence, leveraging generated token likelihoods \citep{guo2017calibrationmodernneuralnetworks,ma2025estimatingllmuncertaintylogits,ni2024large,ni2025knowledge}; and internal state probing confidence, examining hidden states \citep{azaria2023internalstatellmknows,ni2025fullyexploitingllminternal}. 
Differently, our work investigates knowledge boundary perception in LVLMs and provides the first systematic comparison of these methods in the multimodal setting.

\vspace{1em}
\noindent\textbf{LVLMs.} Previous studies have established the widespread adoption of LVLMs in safety-critical domains such as healthcare \citep{NEURIPS2023_5abcdf8e,Hu_2024_CVPR} and autonomous driving \citep{Cui_2024_WACV, jiang2024sennabridginglargevisionlanguage}. While these applications demonstrate LVLMs' functional capabilities, studies show LVLMs frequently produce hallucinations \citep{bai2025hallucinationmultimodallargelanguage,sahoo2024comprehensivesurveyhallucinationlarge}. The current body of work investigates this limitation on different aspects. Some work surveys hallucination types and their causes \citep{liu2024surveyhallucinationlargevisionlanguage,zhou2024analyzingmitigatingobjecthallucination, lan2024surveyhallucinationlargevisual}, while others focus on mitigating hallucinations \citep{li2025mitigatinghallucinationlargevision,wang2024mitigatinghallucinationslargevisionlanguage,Xiao_Huang_Gan_He_Li_Yu_Shu_Jiang_Zhu_2025}. A distinct but less explored research thread investigates LVLMs' knowledge boundary as a potential framework for enhancing model reliability \citep{chen2025detectingknowledgeboundaryvision,wang2024drawinglineenhancingtrustworthiness, Leng_2024_CVPR}. We take this line of work a step further by introducing a novel comparative paradigm that compares perception between LVLMs with their LLM counterparts.

\section{Preliminaries}
\label{sec:pre}
In this section, we provide an overview of our task.

\subsection{Task Formulation}

\noindent \textbf{Visual Question Answering.} The goal of visual question answering (VQA) can be described as follows. For a given question $q$ and an image $i$, the model is asked to provide an answer $a$ based on the question $q$ and image $i$, that is, $a = f_{model}(q, i)$.  

\noindent \textbf{LVLM Knowledge Boundary Perception.} We assess the perception of LVLM's knowledge boundary with the alignment between confidence and 
its actual performance.
Here, we use the model's visual question answering correctness to serve as a proxy for performance and elicit different kinds of model confidence estimates.

\noindent \textbf{Confidence Estimation.}
In this paper, we conduct experiments on the following three kinds of model confidence estimates. As widely adopted and training-free, they can be elicited without changing the internal knowledge of models.

Probabilistic confidence is elicited through the aggregation of token probabilities for scoring, followed by applying a threshold to binarize the score into confidence. It is efficient but only captures lexical-level confidence and requires threshold tuning on a held-out set, which leads to poor generalizability. Some studies also argue that it is not applicable to black-box models \citep{kuhn2023semanticuncertaintylinguisticinvariances}.

Answer consistency-based confidence is elicited by calculating the consistency of multiple generated responses. The core idea is that if the model knows the correct answer, multiple sampled answers should be semantically consistent. It better captures semantics than probabilistic confidence, but is computationally expensive and still requires fitting a threshold \citep{manakul2023selfcheckgptzeroresourceblackboxhallucination}.

Verbalized confidence is elicited by directly asking the model to express confidence \citep{yang2024alignmenthonesty}. Compared to the other two confidences, this confidence reflects models' self-awareness of their knowledge boundaries. Moreover, it eliminates the need for threshold fitting and multiple sampling. Therefore, this kind of confidence receives our primary focus.

\section{Knowledge Boundary Perception in LVLMs}

This section introduces experimental setup to evaluate LVLMs' knowledge boundary perception ability. Along with the elicited confidence and confidence calibration methods evaluated by us.

\subsection{Existing Methods}
\label{sec:methods}
Here, we systematically introduce three basic confidence estimates in Section~\ref{sec:pre}, along with several confidence calibration methods originally designed for LLMs. We also propose new methods. Detailed prompts are in Appendix~\ref{app:prompts}. Basic confidence estimates are in \underline{underline}, and others are existing confidence calibration methods.
\subsubsection{Vanilla Confidence Estimation Methods}

Probabilistic confidence is elicited through token probabilities. Here, we focus on the output perplexity of models.

\noindent\textbullet\enspace\textbf{\underline{Perplexity Threshold (PPL-Thr)}}\hangindent=1em \hangafter=1: Perplexity quantifies a model's uncertainty in content generation \citep{cooper2024perplexedunderstandinglargelanguage}. We binarize this metric into confidence by applying a threshold decided on a held-out set.

Answer consistency-based confidence requires models to generate multiple responses, compute their consistency, and apply a threshold to the consistency scores for confidence elicitation. we implement a two-phase generation protocol: First, generating a reference answer with temperature = 0; Then sampling 10 variant answers with temperature = 1.0, with semantic equivalence between the basic answer and sampled answers evaluated by Qwen2.5-0.5B.

\noindent\textbullet\enspace\textbf{\underline{Random Sample (Random)}}\hangindent=1em \hangafter=1: Simply sample responses without modifying input.

We evaluate two types of verbalized confidence: (1) Single-step verbalized confidence, which is generated simultaneously with the answer, and (2) Double-step verbalized confidence, which is generated by asking the model for an answer in the initial round of dialogue, then providing its confidence in the second round. The distinction between them lies in cognitive focus allocation: single-step confidence elicitation demands concurrent attention to both answer and confidence generation, whereas double-step confidence elicitation enables sequential processing.

\noindent\textbullet\enspace\textbf{\underline{Single-step Vanilla (Vanilla)}}  \hangindent=1em \hangafter=1: Simply ask the model to generate both the answer and confidence in a single interaction.

\noindent\textbullet\enspace\textbf{\underline{Double-step Self Judging (Self-Jud)}}\hangindent=1em \hangafter=1: First, acquiring the model to provide an answer to the question, then asking it to generate confidence.

\subsubsection{Calibrating Verbalized Confidence}

The three methods below aim to calibrate single-step verbalized confidence:

\noindent\textbullet\enspace\textbf{Chain-of-Thought (CoT)}\hangindent=1em \hangafter=1: Zero-shot Chain-of-Thought prompting, Applying ``Analyze step by step'' to the query \citep{kojima2023largelanguagemodelszeroshot}.

\noindent\textbullet\enspace\textbf{Punish}\hangindent=1em \hangafter=1: Penalizing overconfidence via the instruction ``You will be punished if the answer is not right but you say certain''.

\noindent\textbullet\enspace\textbf{Explain}\hangindent=1em \hangafter=1: Requesting models to provide answer explanations before generating their confidence.

The three methods below aim to calibrate double-step verbalized confidence:

\noindent\textbullet\enspace\textbf{Chain-of-Thought (CoT)}\hangindent=1em \hangafter=1: Applying the Chain-of-Thought prompt in the confidence elicitation round of dialogue.

\noindent\textbullet\enspace\textbf{Challenge}\hangindent=1em \hangafter=1: We prepend the critical prompt ``I don't think your answer is right'' to the query in the confidence elicitation round in order to guide the model to be less overconfident.

\noindent\textbullet\enspace\textbf{Punish}\hangindent=1em \hangafter=1: Applying the Punish prompt in the confidence elicitation round of dialogue.

\subsubsection{Calibrating Answer Consistency-Based Confidence}

\noindent\textbullet\enspace\textbf{Rephrasing (Rephr)}\hangindent=1em \hangafter=1: To address persistent errors caused by a specific question phrase, rephrase the original question into semantically equivalent variants with different phrases \citep{yang2024justrephraseituncertainty}. 

\noindent\textbullet\enspace\textbf{Noised Image (Noised-Img)}\hangindent=1em \hangafter=1: Reducing persistent errors caused by a specific image by creating semantically equivalent variants through the addition of subtle noise to the original image. 

\noindent\textbullet\enspace\textbf{Rephrasing and Noised Image (Reph+Nois)}\hangindent=1em \hangafter=1: A combination of the Rephrasing and the Noised Image methods.

\setlength{\fboxsep}{3pt}
\setlength{\fboxrule}{0.4pt}

\begin{table*}[t]
\centering
\small

\caption{Performance of alignment on three datasets and three LVLMs. Best results of each kind of confidence in \textbf{bold} and second best in \underline{underline}. Experimental observations show that LLaVA demonstrates a pattern of complete answer denial when being challenged. We therefore omitted these data from our results.}
\resizebox{\textwidth}{!}{

\begin{tabular}{@{}cccccccccc@{}}
\toprule
\multirow{3}{*}{\textbf{method}} & \multicolumn{3}{c}{\textbf{Qwen2.5-VL}} & \multicolumn{3}{c}{\textbf{LLaVA-1.5}} & \multicolumn{3}{c}{\textbf{DeepSeek-VL2}} \\
\cmidrule(lr){2-4} \cmidrule(lr){5-7} \cmidrule(l){8-10}
& \textbf{Dyn-VQA} & \textbf{Visual7W} & \textbf{MMMU Pro} & \textbf{Dyn-VQA} & \textbf{Visual7W} & \textbf{MMMU Pro} & \textbf{Dyn-VQA} & \textbf{Visual7W} & \textbf{MMMU Pro} \\
\midrule

\colorbox{singlestep}{\makebox[\boxwidth]{\textbf{\underline{Vanilla}}}} & 0.7623 & 0.5840 & 0.4909 & 0.5338 & 0.4140 & 0.2509 & 0.6527 & 0.2820 & 0.2727 \\
\colorbox{singlestep}{\makebox[\boxwidth]{\textbf{CoT}}} & \underline{0.7824} & \underline{0.6080} & \underline{0.6818} & \underline{0.5375} & 0.3940 & 0.2418 & 0.6362 & \underline{0.5540} & \underline{0.3836} \\
\colorbox{singlestep}
{\makebox[\boxwidth]{\textbf{Punish}}} & 0.7112 & 0.5520 & 0.5000 & 0.4899 & \textbf{0.4180} & \textbf{0.3745} & \textbf{0.7093} & 0.3500 & 0.3145 \\
\colorbox{singlestep}{\makebox[\boxwidth]{\textbf{Explain}}} & \textbf{0.8117} & \textbf{0.6180} & 0.5782 & 0.4534 & 0.3900 & 0.2109 & \underline{0.6984} & \textbf{0.5700} & 0.3491 \\
\colorbox{singlestep}{\makebox[\boxwidth]{\textbf{Img-CoT}}} & 0.7276 & 0.6060 & \textbf{0.7182} & \textbf{0.5484} & \underline{0.4140} & \underline{0.2964} & 0.6344 & 0.5360 & \textbf{0.5236} \\ 
\midrule

\colorbox{doublestep}{\makebox[\boxwidth]{\textbf{\underline{Self-Jud}}}} & 0.3272 & 0.5500 & \underline{0.5609} & \underline{0.2468} & \underline{0.4220} & \underline{0.3327} & 0.1993 & 0.4780 & 0.4236 \\
\colorbox{doublestep}{\makebox[\boxwidth]{\textbf{CoT}}} & \underline{0.6435} & \underline{0.5700} & 0.5255 & 0.1463 & 0.4200 & 0.3218 & 0.2029 & 0.4760 & 0.4255 \\
\colorbox{doublestep}{\makebox[\boxwidth]{\textbf{Challenge}}} & \textbf{0.8080} & 0.5280 & 0.4891 & \sout{0.8995} & \sout{0.5800} & \sout{0.6782} & \textbf{0.8007} & 0.5240 & \textbf{0.5709} \\ 
\colorbox{doublestep}{\makebox[\boxwidth]{\textbf{Punish}}} & 0.3272 & 0.5300 & 0.5164 & 0.1298 & 0.4200 & 0.3218 & 0.4936 & \underline{0.5300} & 0.4345 \\
\colorbox{doublestep}{\makebox[\boxwidth]{\textbf{Prob-Thr}}} & 0.5960 & \textbf{0.5820} & \textbf{0.5855} & \textbf{0.7971} & \textbf{0.6140} & \textbf{0.6091} & \underline{0.6910} & \textbf{0.6060} & \underline{0.5218} \\
\midrule

\colorbox{consistency}{\makebox[\boxwidth]{\textbf{\underline{Random}}}} & 0.5448 & 0.5700 & 0.5327 & \underline{0.8976} & \textbf{0.7080} & \textbf{0.6709} & 0.8026 & \underline{0.6460} & \textbf{0.6000} \\
\colorbox{consistency}{\makebox[\boxwidth]{\textbf{Noised Img}}} & 0.7313 & \underline{0.6000} & 0.5400 & 0.8958 & 0.6740 & 0.6655 & 0.8062 & 0.6300 & \underline{0.5818} \\
\colorbox{consistency}{\makebox[\boxwidth]{\textbf{Rephr}}} & \underline{0.8026} & 0.5660 & 0.5364 & \underline{0.8976} & \underline{0.6920} & \underline{0.6672} & \underline{0.8080} & 0.6260 & 0.5764 \\
\colorbox{consistency}{\makebox[\boxwidth]{\textbf{Reph+Nois}}} & 0.7733 & 0.5500 & \underline{0.5509} & \textbf{0.9013} & 0.6780 & 0.6655 & \textbf{0.8099} & 0.6120 & 0.5618 \\
\colorbox{consistency}{\makebox[\boxwidth]{\textbf{{Cross Model}}}} & \textbf{0.8208} & \textbf{0.6320} & \textbf{0.5800} & \underline{0.8976} & 0.6520 & 0.6618 & 0.8062 & \textbf{0.6740} & 0.5964 \\
\midrule

\colorbox{PPL}{\makebox[\boxwidth]{\textbf{\underline{PPL Thr}}}} & 0.7916 & 0.6020 & 0.6073 & 0.8519 & 0.7060 & 0.6800 & 0.7934 & 0.6280 & 0.5345 \\
\bottomrule
\end{tabular}}
\vspace{-0.5em}
\label{tab:align}
\end{table*}

\subsection{Newly Proposed Methods}

\noindent\textbullet\enspace\textbf{Image Chain of Thought (Img-CoT)}\hangindent=1em \hangafter=1: Prompting models to generate textual image descriptions before reasoning to convert visual modality information to textual modality. 

\noindent\textbullet\enspace\textbf{Probability Threshold (Prob-Thr)}\hangindent=1em \hangafter=1: Prompting models to generate continuous probabilities of answers (0–1), then applies a threshold to them to generate binary confidences. The threshold is decided on a held-out set. 

\noindent\textbullet\enspace\textbf{Cross Model}\hangindent=1em \hangafter=1: Utilizing generated responses from different models to calculate the consistency score. We generate answers using the three models mentioned in the next subsection. The primary model generates four responses, while the other two models each generate three responses. We then calculate their consistency with the answer obtained through greedy sampling from the primary model to derive the confidence. This method can be viewed as using other models' answers to evaluate whether the answer generated by a given model is reliable.

\subsection{Experimental Setup}
\noindent\textbf{Datasets.} We conduct experiments on three VQA benchmark datasets. They emphasize on LVLM's different abilities. Visual7W \citep{Zhu_2016_CVPR} emphasizes abilities in vision comprehension, it contains 70K image-QA pairs for basic visual understanding. Dyn-VQA \citep{li2025benchmarkingmultimodalretrievalaugmented} emphasizes language reasoning, it contains 1.5K questions testing multi-modal knowledge and multi-hop reasoning.; MMMU Pro \citep{yue2024mmmuprorobustmultidisciplinemultimodal} emphasizes both vision and language capability, it contains 12K expert-curated multimodal questions. For evaluation, we respectively sample 550 questions from Dyn-VQA and MMMU Pro datasets, and sample 500 questions from the Visual7W dataset.

\noindent\textbf{Models. } We conduct experiments on three representative LVLMs: Qwen2.5-VL-7B \citep{bai2025qwen25vltechnicalreport}, DeepSeek-VL2-16B \citep{wu2024deepseekvl2mixtureofexpertsvisionlanguagemodels}, and LLaVA-v1.5-7B \citep{liu2024improvedbaselinesvisualinstruction}.
We selected these three LVLMs because they are widely adopted and serve as established baselines in the field. Additionally, since all three models are constructed by integrating visual encoders with their corresponding LLMs, this choice enables a parallel comparison between the performance of these LVLMs and their respective LLMs in subsequent analyses.

\begin{figure}[ht] 
\centering
\includegraphics[width=0.5\linewidth]{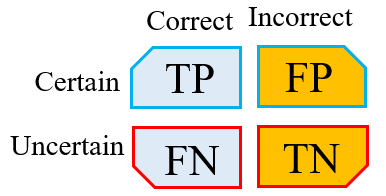}
\caption{Count of samples for various matches between
answer correctness and model confidence. We use $Total=FN+FP+TN+TP$ to represent the total number of samples.}
\vspace{-1em}
\end{figure}
\sloppy
\noindent\textbf{Metrics. } We mainly utilize the evaluation metrics proposed by \citet{ni2024llmsneedretrievalaugmentation}: 
\fussy
(1) \textbf{Uncertain-Rate (Unc-R.)}: $\frac{FN+TN}{Total}$ represents the proportion where the judgement of the answer is unconfident.
(2) \textbf{Accuracy (Acc.)}: $\frac{TP+FN}{Total}$ indicates the ratio of correct answers generated by the model.
(3) \textbf{Alignment (Align.)}: $\frac{TP+TN}{Total}$ represents the proportion of samples where confidence matches the result, we mainly use this metric to assess the model's knowledge boundary perception ability.
(4) \textbf{Overconfidence (Overco.)}: $\frac{FP}{Total}$ is the ratio of model-generated answer is incorrect, but the judgement is confident.
(5) \textbf{Conservativeness (Conser.)}: $\frac{FN}{Total}$ is the ratio of model-generated answer is correct but the judgement is unconfident.

\subsection{Results and Analysis}
\label{sec:LVLM performance}
Table~\ref{tab:align} presents the results of alignment performance across different datasets and models. Please refer to Appendix~\ref{app:LVLM} for implementation details and detailed results.

\subsubsection{Performance of Different Types of Confidence}
Here, we analyze three basic elicited confidence's performance. Our findings are as follows: 

1) \textbf{Compared to verbalized and probabilistic confidence, answer consistency-based confidence often shows higher alignment.} As shown in Table~\ref{tab:align}, the basic answer-consistency based confidence (Random) achieves higher alignment compared to verbalized (Vanilla, Self-Jud) and probabilistic confidences (PPL Thr) on both LLaVA-1.5 and Deepseek-VL2. This may be because, unlike probabilistic confidence that operates at the lexical level, answer consistency-based confidence better captures semantics by evaluating answer consistency \citep{kuhn2023semanticuncertaintylinguisticinvariances}, achieving higher alignment. Additionally, while verbalized confidence is uncalibrated, eliciting answer consistency-based confidence calibrating a threshold on a held-out set, further improves alignment.

Despite answer consistency-based confidence exhibiting high alignment, it comes at a cost: eliciting this kind of confidence requires generating multiple responses, incurring high computational costs. And its reliance on a held-out set for threshold calibration limits its generalizability.

2) \textbf{Probabilistic confidence surpasses verbalized confidence in alignment performance.} 
As shown in Table~\ref{tab:align}, probabilistic confidence's alignment performance consistently surpasses verbalized confidence, and it outperforms answer consiste\-ncy-based confidence on Qwen2.5-VL. Though it falls behind consistency-based confidence on LLAVA-1.5 and DeepSeek-VL2, the alignment differences are small. Additionally, it functions more efficiently without the high computational cost of generating multiple responses. 

However, probabilistic confidence, like answer consistency-based confidence, still requires threshold calibration on a held-out set, which affects its generalizability.

\begin{table}[t]
\centering
\small
\caption{The performance of verbalized confidence on Qwen2.5-VL, single-step confidences are in \colorbox{singlestep}{blue} and double-step confidences are in \colorbox{doublestep}{orange}.}
\resizebox{0.48\textwidth}{!}{
\begin{tabular}{@{}c|cc|cc|cc|@{}}
\toprule
\multirow{4}{*}{\textbf{method}} & \multicolumn{2}{c}{\textbf{Dyn-VQA}} & \multicolumn{2}{c}{\textbf{Visual7W}} & \multicolumn{2}{c}{\textbf{MMMU Pro}} \\
\cmidrule(lr){2-3} \cmidrule(lr){4-5} \cmidrule(l){6-7}
& \textbf{Conser.}  & \textbf{Overco.}& \textbf{Conser.}  & \textbf{Overco.} & \textbf{Conser.} &  \textbf{Overco.} \\
\midrule

\colorbox{singlestep}{\makebox[\boxwidthh]{\textbf{Vanilla}}} & \cellcolor[RGB]{253,236,226} 0.1024 & \cellcolor[RGB]{230,239,248} 0.1353 & \cellcolor[RGB]{255,255,254} 0.0900 & \cellcolor[RGB]{249,251,253}0.3260 & \cellcolor[RGB]{255,251,249}0.1327 & \cellcolor[RGB]{255,252,249}0.3764 \\
\colorbox{singlestep}{\makebox[\boxwidthh]{\textbf{CoT}}} & \cellcolor[RGB]{253,254,254}0.0786 & \cellcolor[RGB]{232,241,249}0.1389 & \cellcolor[RGB]{254,245,240}0.1340 & \cellcolor[RGB]{228,239,248}0.2580 & \cellcolor[RGB]{255,255,255}0.1127 & \cellcolor[RGB]{209,227,243}0.2055 \\
\colorbox{singlestep}{\makebox[\boxwidthh]{\textbf{Punish}}} & \cellcolor[RGB]{255,255,254}0.0804 & \cellcolor[RGB]{255,250,247}0.2084 & \cellcolor[RGB]{249,251,253}0.0820 & \cellcolor[RGB]{254,241,234}0.3660 & \cellcolor[RGB]{255,255,255}0.1127 & \cellcolor[RGB]{254,242,235}0.3873 \\
\midrule

\colorbox{doublestep}{\makebox[\boxwidthh]{\textbf{Self-Jud}}} & \cellcolor[RGB]{155,194,230}0.0018 & \cellcolor[RGB]{244,176,132}0.6709 & \cellcolor[RGB]{163,198,232}0.0180 & \cellcolor[RGB]{247,195,161}0.4320 & \cellcolor[RGB]{209,226,243}0.0701 & \cellcolor[RGB]{254,254,254}0.3692 \\
\colorbox{doublestep}{\makebox[\boxwidthh]{\textbf{CoT}}}& \cellcolor[RGB]{195,218,240}0.0329 & \cellcolor[RGB]{252,232,218}0.3236 & \cellcolor[RGB]{176,207,235}0.0280 & \cellcolor[RGB]{250,216,194}0.4020 & \cellcolor[RGB]{241,246,251}0.1000 & \cellcolor[RGB]{255,253,252
}0.3745 \\
\colorbox{doublestep}{\makebox[\boxwidthh]{\textbf{Punish}}} & \cellcolor[RGB]{155,194,230}0.0018 & \cellcolor[RGB]{244,176,132}0.6709 & \cellcolor[RGB]{155,194,230}0.0120 & \cellcolor[RGB]{244,176,132}0.4580 & \cellcolor[RGB]{244,176,132}0.0200 & \cellcolor[RGB]{244,176,132}0.4636 \\
\bottomrule
\end{tabular}}

\vspace{-1.5em}
\label{tab:verbalization}
\end{table}

\begin{table*}[t]
\centering
\small
\caption{The performance of single-step reasoning elicitation methods on Qwen2.5-VL.}
\resizebox{\textwidth}{!}{
\begin{tabular}{@{}cccccccccc@{}}
\toprule
\multirow{4}{*}{\textbf{method}} & \multicolumn{3}{c}{\textbf{Dyn-VQA}} & \multicolumn{3}{c}{\textbf{Visual7W}} & \multicolumn{3}{c}{\textbf{MMMU Pro}} \\
\cmidrule(lr){2-4} \cmidrule(lr){5-7} \cmidrule(l){8-10}
& \textbf{Acc.} & \textbf{Align.} & \textbf{Overco.} & \textbf{Acc.} & \textbf{Align.} & \textbf{Overco.} & \textbf{Acc.} & \textbf{Align.} & \textbf{Overco.} \\
\midrule
\colorbox{singlestep}{\makebox[\boxwidth]{\textbf{Vanilla}}} & 0.1846 & 0.7623 & 0.1353 & 0.4380 & 0.5840 & \textbf{0.3260} & 0.4564 & 0.4909 & \textbf{0.3764} \\
\colorbox{singlestep}{\makebox[\boxwidth]{\textbf{CoT}}} & \textbf{0.2121} & \underline{0.7824} & \underline{0.1389} & \underline{0.4920} & \underline{0.6080} & 0.2580 & \underline{0.6436} & \underline{0.6818} & 0.2055 \\
\colorbox{singlestep}{\makebox[\boxwidth]{\textbf{Img-CoT}}} & \underline{0.2048} & 0.7276 & \textbf{0.2066} & \textbf{0.5020} & 0.6060 & \underline{0.3080} & \textbf{0.6636} & \textbf{0.7182} & 0.1691 \\
\colorbox{singlestep}{\makebox[\boxwidth]{\textbf{Explain}}} & 0.1956 & \textbf{0.8117} & 0.0823 & 0.4740 & \textbf{0.6180} & 0.2720 & 0.5309 & 0.5782 & \underline{0.2982} \\


\bottomrule
\end{tabular}}

\label{tab:reason}
\end{table*}

3) \textbf{Verbalized confidence demonstrates lower alignment compared to probabilistic and answer consistency-based confidences, and judges answers overconfidently.} Compared to probabilistic and answer consistency-based confidences, eliciting verbalized confidence is computationally efficient and generalizable. However, as shown in Table~\ref{tab:align}, both single-step (Vanilla) and double-step (Self-Jud) verbalized confidences' alignment are lower than the other two confidences. To investigate the cause of it, we calculate the conservativeness and overconfidence on verbalized confidence, as shown in Table~\ref{tab:verbalization}, we find that the ratio of overconfident responses is substantially higher than conservative responses. 
This pattern suggests that LVLMs, like LLMs, are intrinsically biased toward affirming their own output \citep{groot2024overconfidencekeyverbalizeduncertainty, sun2025largelanguagemodelsoverconfident}.

Table~\ref{tab:verbalization} also shows that double-step verbalized confidence exhibits more severe overconfidence than its single-step counterpart. This may be because the model's self-generated answers in the first round of dialogue serve as false positive signals of its capability, reinforcing overconfident behavior through misleading the model to self-affirmation.

\subsubsection{Confidence Calibration in LVLMs}
In this section, we evaluate the effectiveness of existing confidence calibration methods developed for LLMs in the context of LVLMs, as well as our proposed methods.

For existing confidence calibration methods, our observations are as follows:

1) \textbf{Single-step reasoning elicitation methods effectively enhance the accuracy and alignment of LVLMs.} As shown in Table~\ref{tab:align}, we found reasoning elicitation methods (Explain, CoT, and Img-CoT) exhibit high alignment. To further investigate them, we calculate other metrics about them. Table~\ref{tab:reason} shows that different reasoning elicitation methods excel on specific datasets: CoT method improves alignment and accuracy across all datasets and causes overconfidence on Dyn-VQA. The Explain method outperforms CoT in alignment on Visual7W and Dyn-VQA datasets. This observed difference may stem from the Explain method's design: while the CoT method enforces step-by-step reasoning, the Explain method prioritizes direct justification, thus reducing redundant context for simple questions and improving the calibration of LVLMs' confidence outputs.

2) \textbf{Answer consistency-based confidence calibration methods improve alignment on Qwen2.5-VL, but show limited effectiveness on other models.} We observed that, even when sampling responses at the same temperature of 1.0, models differ in their output diversity. As shown in Table~\ref{tab:align}, when random sampling Qwen2.5-VL's responses, it tends to generate consistent yet incorrect responses, resulting in low alignment. However, both the rephrasing and the noised image methods show effectiveness in mitigating this tendency, consequently achieving higher alignment. In contrast, LLaVA-1.5 and DeepSeek-VL2 generate more diverse outputs when the response is incorrect, allowing the Random Sampling method to perform well and making Noised Image and Rephrasing methods less effective in enhancing alignment by comparison.

We propose Image Chain of Thought, Probability Threshold, and Cross Model Consistency methods in Section~\ref{sec:methods}, their performances are as follows:

\begin{table*}[t]
\centering
\small
\caption{LLMs and LVLMs comparison for single-step verbalization based methods on Dyn-VQA.}
\resizebox{\textwidth}{!}{
\begin{tabular}{@{}ccccccc|ccccc|ccccc|@{}}
\toprule

\multirow{4}{*}{\textbf{method}} & \multirow{4}{*}{\textbf{Model}} & \multicolumn{5}{c}{Qwen2.5} & \multicolumn{5}{c}{DeepSeek-VL2} & \multicolumn{5}{c}{LLaVA-1.5}\\
\cmidrule(lr){3-7} \cmidrule(lr){8-12} \cmidrule(l){13-17}
& &  \textbf{Unc-R.} & \textbf{Acc} & \textbf{Align.} & \textbf{Conser.} & \textbf{Overco.} & \textbf{Unc-R.} & \textbf{Acc} & \textbf{Align.} & \textbf{Conser.} & \textbf{Overco.} & \textbf{Unc-R.} & \textbf{Acc} & \textbf{Align.} & \textbf{Conser.} & \textbf{Overco.} \\
\midrule
\multirow{2}{*}{\colorbox{singlestep}{\makebox[\boxwidthh]{\textbf{Vanilla}}}} & LVLM & 0.782 & 0.185 & \textbf{0.762} & 0.102 & \textbf{0.135} & \textbf{0.788} & 0.146 & \textbf{0.653} & \textbf{0.141} & 0.207 & \textbf{0.490} & 0.088 & \textbf{0.534} & \textbf{0.022} & 0.444 \\
& LLM & \textbf{0.788} & \textbf{0.285} & 0.729 & \textbf{0.172} & 0.099 & 0.161 & \textbf{0.225} & 0.338 & 0.024 & \textbf{0.638} & 0.011 & \textbf{0.141} & 0.152 & 0.001 & \textbf{0.848} \\
\addlinespace[0.1em]
\cdashline{2-17}[3.6pt/2pt]
\addlinespace[0.3em]
\multirow{2}{*}{\colorbox{singlestep}{\makebox[\boxwidthh]{\textbf{CoT}}}} & LVLM & 0.728 & 0.212 & \textbf{0.782} & \textbf{0.079} & 0.139 & \textbf{0.638} & 0.170 & \textbf{0.636} & \textbf{0.086} & 0.278 & \textbf{0.512} & 0.084 & \textbf{0.538} & \textbf{0.031} & 0.431 \\
& LLM & 0.448 & \textbf{0.294} & 0.651 & 0.046 & \textbf{0.304} & 0.095 & \textbf{0.296} & 0.362 & 0.015 & \textbf{0.623} & 0.117 & \textbf{0.199} & 0.302 & 0.007 & \textbf{0.691} \\
\addlinespace[0.1em]
\cdashline{2-17}[3.6pt/2pt]
\addlinespace[0.3em]
\multirow{2}{*}{\colorbox{singlestep}{\makebox[\boxwidthh]{\textbf{Punish}}}} & LVLM & 0.711 & 0.168 & \textbf{0.711} & 0.080 & \textbf{0.208} & \textbf{0.848} & 0.161 & \textbf{0.709} & \textbf{0.150} & 0.141 & \textbf{0.450} & 0.095 & \textbf{0.490} & \textbf{0.027} & 0.483 \\
& LLM & \textbf{0.956} & \textbf{0.294} & 0.713 & \textbf{0.269} & 0.018 & 0.266 & \textbf{0.229} & 0.455 & 0.020 & \textbf{0.525} & 0.057 & \textbf{0.152} & 0.201 & 0.004 & \textbf{0.795} \\
\addlinespace[0.1em]
\cdashline{2-17}[3.6pt/2pt]
\addlinespace[0.3em]
\multirow{2}{*}{\colorbox{singlestep}{\makebox[\boxwidthh]{\textbf{Explain}}}} & LVLM & \textbf{0.828} & 0.196 & \textbf{0.812} & \textbf{0.106} & 0.082 & \textbf{0.786} & 0.168 & \textbf{0.698} & \textbf{0.128} & 0.174 & \textbf{0.421} & 0.084 & \textbf{0.453} & \textbf{0.027} & 0.519 \\
& LLM & 0.536 & \textbf{0.298} & 0.673 & 0.080 & \textbf{0.247} & 0.079 & \textbf{0.252} & 0.320 & 0.006 & \textbf{0.675} & 0.159 & \textbf{0.219} & 0.364 & 0.007 & \textbf{0.629} \\
\bottomrule
\end{tabular}}

\label{tab:LLM vs LVLM}
\end{table*}

1) \textbf{Image Chain of Thought method effectively enhances alignment and accuracy on MMMU Pro. }As shown in Table~\ref{tab:reason}, Img-CoT demonstrates remarkable performance on the MMMU Pro dataset, which requires both strong visual perception and reasoning capabilities. It improves accuracy and alignment, outperforming CoT method. This indicates that its mechanism for converting visual modality into language modality can effectively enhance models' comprehension of the content in the image, thereby achieving superior performance. 
However, it fails to improve alignment on Dyn-VQA and Visual7W datasets, as their images lack complex objects like sheet music or circuit diagrams which MMMU Pro contains. The forced "describe the image" process may lead to excessive descriptions, creating false positives in capability assessment and increasing overconfidence. You can refer to Appendix 4 for typical cases where Img-CoT makes the model overconfident, while the CoT method does not.

2) \textbf{Probability Threshold method shows higher alignment than other double-step verbalizated confidence calibration methods. } As shown in Table~\ref{tab:align}, the Probability Threshold method outperforms alternative double-step methods. Despite the need to calibrate the threshold, it effectively enhances alignment.

3) \textbf{Cross-Model Method Enhances Alignment for Qwen2.5-VL}
As demonstrated in Table~\ref{tab:align}, the Cross-Model method significantly outperforms other answer consistency-based confidence calibration approaches for Qwen2.5-VL. Our results reveals that under random sampling conditions, Qwen2.5-VL shows weaker alignment between its consistency scores and actual capabilities across all three datasets compared to DeepSeek-VL2 and LLaVA-v1.5, which maintain stronger alignment. The Cross-Model approach addresses this limitation by incorporating responses from these better-aligned models, thereby improving the confidence calibration and capability alignment for Qwen2.5-VL.
\vspace{-1em}

\begin{figure*}[t]
  \centering
   \begin{subfigure}[b]{0.32\textwidth}
    \includegraphics[width=\textwidth]{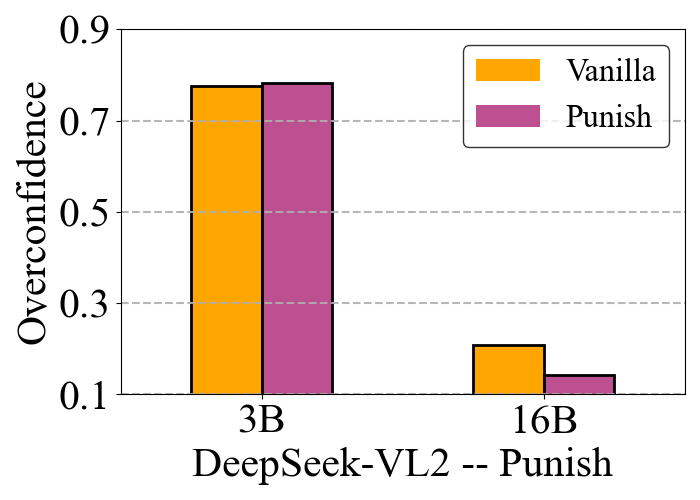}
  \end{subfigure}
  \hfill
  \begin{subfigure}[b]{0.32\textwidth}
    \includegraphics[width=\textwidth]{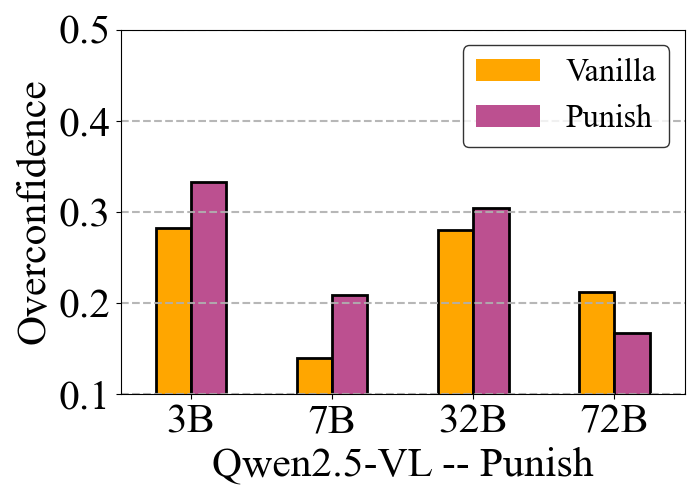}
  \end{subfigure}
  \hfill
  \begin{subfigure}[b]{0.32\textwidth}
    \includegraphics[width=\textwidth]{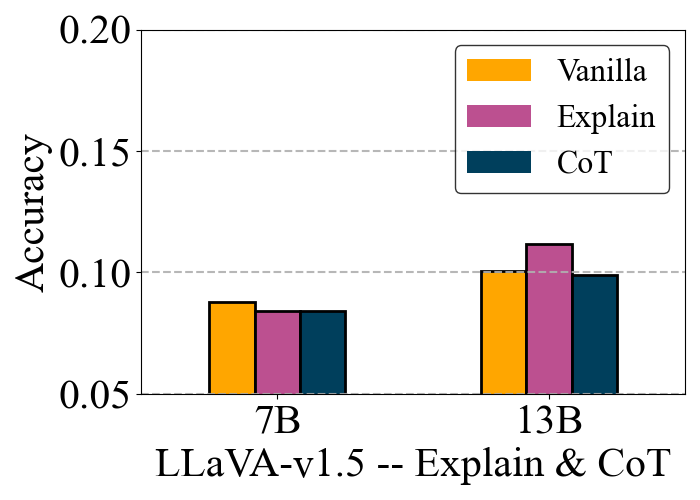}
  \end{subfigure}
  \hfill
    \caption{Comparative analysis of instruction following ability across model scales.}
    \label{fig:scale}
  \smallskip
  
\end{figure*}

\section{Perception Comparison Between LVLMs and LLMs}
\label{sec:LLMs vs LVLMs}
Compared to LLMs, LVLMs need to process additional visual modality and integrate information across different modalities. This raises a question: how does the perception of LVLMs differ from that of LLMs? Knowing these distinctions is valuable for developing trustworthy LVLMs.

In this section, we investigate the difference of knowledge boundary perception between LVLMs and their LLM counterparts. Focusing on verbalized confidence cause it directly reflects models' self-awareness of their knowledge boundaries.
We further propose several hypotheses about these differences' underlying causes and validate them through the comparison between different model scales and input modalities.

\subsection{Experimental Setup}
\noindent\textbf{Datasets. }In this section, we mainly focus on Dyn-VQA dataset. Dyn-VQA provides both VQA question image pairs and their semantically equivalent QA questions (e.g., QA: ``How many humans have landed on Mars?'' vs. VQA: ``How many humans have landed on this planet?'' with an image of Mars). This enables fair 
model performance comparison across text-only modality and vision-text modality inputs.

\noindent\textbf{Models. }In this section, we compare LVLMs with their base LLM counterparts to ensure fair comparison: Qwen2.5-VL, DeepSeek-VL2, LLaVA-v1.5 vs Qwen2.5, DeepSeek-MoE, Vicuna-v1.5.

\subsection{Results and Analysis}
Here, we apply VQA queries on LVLMs, and their semantic equivalent QA queries on LLMs to fairly compare them. And focus on single-step verbalized confidence. We defer results about other kinds of confidence to Appendix~\ref{app:compare}. Here are our findings:

1) \textbf{Compared to LLMs, LVLMs struggle to follow certain methods' instructions, leading to performance deviating from expected.} As shown in Table~\ref{tab:LLM vs LVLM}, Qwen2.5-VL cannot effectively follow the Punish instruction. As a result, this method not only fails to reduce overconfidence but actually exacerbates it, leading to lower alignment than Vanilla. Similarly, LLaVA-1.5 disregards CoT and Explain instructions, persistently generating responses without proper reasoning or explanation, which results in lower accuracy. This stands in contrast to LLMs, where the Punish method effectively reduces Qwen2.5's overconfidence; CoT and Explain instructions reliably ignite reasoning responses in Vicuna-1.5, thus improving its accuracy.

2) \textbf{For single-step verbalized confidence, LVL\-Ms tend to have lower accuracy compared to LLMs. Along with higher alignment due to reduced overconfidence.}
As shown in Table~\ref{tab:LLM vs LVLM}, under all single-step verbalized confidence for the three series of models, the answer accuracy of LVLMs is lower than that of LLMs. Meanwhile, LVLMs exhibit a higher uncertain-rate compared to LLMs. Specifically, LLaVA exhibits an average accuracy reduction of 0.09 with a concurrent 0.382 increase in uncertain-rate than its counterpart LLM. And in DeepSeek-VL2, we observe a 0.089 accuracy decrement paired with a 0.615 surge in uncertainty than LLM. Compared to LLMs, LVLMs' accuracy drop is relatively smaller than their uncertain-rate increase, thus they demonstrate less severe overconfidence than LLMs, leading to relatively higher alignment in their responses.

\subsection{Analysis Across Model Scales and Modalities}

Building upon the findings discussed in the previous subsection, we observe notable performance distinctions between LLMs and LVLMs, which motivate us to propose the following hypothesis regarding their potential underlying causes:


1) Model capacity bottleneck: We hypothesize that the inferior instruction-following abilities of LVLMs stems from their internal capacity limitations, where visual modality integration competes for models' internal parameter resources that would otherwise support language processing capabilities.


2) Cross-modal limitation awareness: While the LVLMs demonstrate lower accuracy than LLMs, their verbalized confidence shows better alignment with performance. We hypothesize this stems from two factors: (1) LVLMs' constrained cross-modal processing ability leads to degraded multimodal VQA accuracy, and (2) LVLMs' awareness of this limitation results in higher alignment.

To validate our capacity hypothesis of instruction following ability, we conduct a comparative analysis on different scale models and find that:

\textbf{As LVLMs scale up, they generally exhibit stronger instruction following capabilities.}
As shown in Figure~\ref{fig:scale}.
For Qwen2.5-VL and DeepSeek-VL2, the Punish method effectively reduces overconfidence in larger models (Qwen2.5-VL-72B, DeepSeek-Vl2-16B) but shows limited impact on smaller ones ( < 32B Qwen2.5-VL, DeepSeek-VL2-3B).
For LLaVA-1.5, the 13B model follows Explain instruction which 7B model not follows, thus Explain improves accuracy in the 13B model.

These phenomena supports our hypothesis: the parameter constraints of small scale LVLMs create a dilemma between visual processing and linguistic comprehension, resulting in degraded language understanding and consequently weaker instruction following ability. In contrast, larger LVLMs allocate more parameters to language processing, maintaining strong language ability while handling multimodal inputs, thus demonstrating stronger instruction following ability.






\begin{table}[t]
\centering
\small
\caption{The performance of LVLMs under different query modalities, we add text question at the bottom of the image to generate pure image ``V''QA query.}
\resizebox{0.48\textwidth}{!}{
\begin{tabular}{@{}ccccccc@{}}
\toprule
\multicolumn{1}{c}{\multirow{4}{*}{\textbf{Model}}} & \multirow{4}{*}{\textbf{Task}} & \multicolumn{5}{c}{\textbf{Dyn-VQA}} \\
\cmidrule(l){3-7}
& & \textbf{Unc-R.} & \textbf{Acc} & \textbf{Align.} & \textbf{Conser.} & \textbf{Overco.} \\
\midrule
\multicolumn{1}{c}{\multirow{3}{*}{\textbf{Qwen2.5-VL}}} & \textbf{``V''QA} & 0.461 & 0.223 & 0.578 & 0.053 & \textbf{0.369} \\
&\textbf{VQA} & \textbf{0.782} & 0.185 & \textbf{0.762} & 0.102 & 0.135 \\
&\textbf{QA} & 0.766 & \textbf{0.252} & 0.700 & \textbf{0.159} & 0.141 \\
\midrule
\multicolumn{1}{c}{\multirow{3}{*}{\textbf{DeepSeek-VL2}}} & \textbf{``V''QA} & 0.227 & 0.208 & 0.435 & 0.000 & \textbf{0.565} \\
&\textbf{VQA} & \textbf{0.788} & 0.146 & \textbf{0.653} & \textbf{0.141} & 0.207 \\
&\textbf{QA} & 0.545 & \textbf{0.256} & 0.559 & 0.121 & 0.320 \\

\bottomrule
\end{tabular}}

\label{tab:modalityquery}
\vspace{-1em} 
\end{table}

To validate our accuracy and alignment hypothesis, we conduct comparative analysis on text-only QA, vision-text VQA, and vision-only "V"QA mod\-ality of queries on LVLMs, our results reveal that: 

\textbf{LVLMs exhibit lower accuracy but higher alignment when responding to multimodal VQA queries.}
As shown in Table~\ref{tab:modalityquery}, both models demonstrate lower accuracy when answering VQA queries that demand cross-modal understanding ability compared to pure text QA and pure image ``V''QA queries. Concurrently, they demonstrate increased uncertain-rate and improved confidence performan\-ce alignment for these multimodal queries.


These observations support our hypothesis:

1. Limited cross-modal ability: LVLMs struggle to effectively synthesize information across modalities, leading to reduced answering accuracy on multimodal queries compared to unimodal queries.

2. Capability awareness: When encountering challenging multimodal queries, LVLMs exhibit self-awareness of their limited ability through generating more uncertainty responses. This decreases overconfidence and thus improves alignment.


\section{Conclusion}
In this paper, we present a systematic investigation of knowledge boundary perception in LVLMs, assessing this ability through alignment. First, we evaluate three kinds of confidence, and observe that answer consistency-based confidence reaches the highest alignment, whereas verbalized confidence induces overconfidence. We also evaluate several confidence calibration methods, with our results revealing that reasoning elicitation methods improve accuracy and alignment, while our proposed methods show effectiveness. Second, we compare LVLMs with LLMs, and reveal that while LVLMs exhibit lower QA accuracy, they achieve higher alignment, which is attributable to LVLMs' awareness of their multimodal integration ability limitation. We 
also observe that LVLMs have weaker instruction following ability than LLMs. 

\section*{Limitations}
First, due to dataset constraints, we only compared LVLMs and LLMs on Dyn-VQA; broader benchmarks are needed for future validation. Second, our analysis did not examine internal model states, leaving internal mechanistic differences in knowledge boundary perception underexplored. Third, we focused on binary confidence measures; extending this to continuous confidence scales could yield finer-grained insights. These limitations highlight directions for future work on LVLM evaluation and interpretability.

\section*{Ethics Statement}
In this paper, all the datasets we use are open-source, and the models we employ are either open-source or widely used. Furthermore, the methods we propose do not induce the model to output any harmful information.

\section*{Acknowledgements}
This work was funded by the National Natural Science Foundation of China (NSFC) under Grant No. 62302486, the Innovation Project of ICT CAS under Grant No. E361140, and the CAS Special Research Assistant Funding Project.

\bibliography{anthology,emnlp2023}
\bibliographystyle{acl_natbib}

\newpage
\appendix

\section{Appendix}
\label{sec:appendix}

\subsection{Prompts}
\label{app:prompts}
\subsubsection{Single Step Verbalization Based Prompts}

\textbf{Vanilla.  } \textit{Answer the question based on your internal knowledge and the image. If you are sure the answer is accurate and correct, please say ``certain'' after the answer. If you are not confident with the answer, please say ``uncertain''.}

\noindent\textit{Question: [Question]}

\noindent\textit{Answer:}

\vspace{1em}
\noindent\textbf{CoT.  } \textit{Answer the question based on your internal knowledge and the image. \textbf{\underline{Analyse step by step}}. If you are sure the answer is accurate and correct, please say ``certain'' after the answer. If you are not confident with the answer, please say ``uncertain''.}

\noindent\textit{Question: [Question]}

\noindent\textit{Answer:}

\vspace{1em}
\noindent\textbf{Img-CoT.  } \textit{Answer the question based on your internal knowledge and the image. \textbf{\underline{First, describe the}}
\textbf{\underline{image, then analyse step by step}}. If you are sure the answer is accurate and correct, please say ``certain'' after the answer. If you are not confident with the answer, please say ``uncertain''.}

\noindent\textit{Question: [Question]}

\noindent\textit{Answer:}

\vspace{1em}
\noindent\textbf{Punish.  } \textit{Answer the question based on your internal knowledge and the image. If you are sure the answer is accurate and correct, please say ``certain'' after the answer. If you are not confident with the answer, please say ``uncertain''. \textbf{\underline{You will be punish-}}
\textbf{\underline{ed if the answer is not right but you say ``certain''.}}}

\noindent\textit{Question: [Question]}

\noindent\textit{Answer:}

\vspace{1em}
\noindent\textbf{Explain.  } \textit{Answer the question based on your internal knowledge and the image, \textbf{\underline{explain why you give}} \textbf{\underline{this answer.}} If you are sure the answer is accurate and correct, please say ``certain'' after the answer. If you are not confident with the answer, please say ``uncertain''.}

\noindent\textit{Question: [Question]}

\noindent\textit{Answer:}

\subsubsection{Double Step Verbalization Based Prompts}
For the double step verbalization based methods, we first prompt the model to generate answer, then prompt the model to give its confidence in the second round chat.

\vspace{1em}
\noindent\textbf{First Round Answer Generation. }\textit{Answer the question based on your internal knowledge and the image.}

\noindent\textit{Question: [Question]}

\noindent\textit{Answer:}

\vspace{1em}
\noindent\textbf{Self-Judging. } \textit{If you are sure your previous answer is accurate and correct, please say ``certain'', If you are not confident with the answer, please say ``uncertain''.}

\vspace{1em}
\noindent\textbf{CoT. } \textit{If you are sure your previous answer is accurate and correct, please say ``certain'', If you are not confident with the answer, please say ``uncertain''. \textbf{\underline{Analyse step by step}}, then provide Your judgement.}

\vspace{1em}
\noindent\textbf{Challenge. } \textit{\textbf{\underline{I don't think your answer is right}}, if you still think your answer is right, please say ``ceratin''. Otherwise, say ``uncertain''.}

\vspace{1em}
\noindent\textbf{Punish. } \textit{If you are sure your previous answer is accurate and correct, please say ``certain'', If you are not confident with the answer, please say ``uncertain''. \textbf{\underline{You will be punished if the answer is not}}\\
\textbf{\underline{right but you say ``certain''.}}}

\vspace{1em}
\noindent\textbf{Probability+Threshold. } \textit{Provide the probability that your answer is correct (0.0 to 1.0). Give ONLY the probability, no other words or explanation.}
\vspace{1em}

\subsubsection{Answer Consistency Based Prompts}

\vspace{1em}
\noindent\textbf{Rephrasing. } \textit{Based on the Following question, generate [number of semantical equivalent questions] semantically equivalent questions. your output should be a list of strings and add a sequnce number with a dot at the start of each output question, like [1.``question1'',2.``question2'',...].}

\noindent\textit{Question: [The original question]}

\noindent\textit{Semantically equivalent questions:}

\newpage
\subsection{LVLMs' Knowledge Boundary Perception Ability}
\label{app:LVLM}

\subsubsection{Implementation Details}
In this section, we provide a detailed introduction to our implementation details.

For content generation, we mainly utilize APIs to generate answers.

For verbalization based methods, we set the model temperature to 0 and set a fixed seed to obtain high-quality and relatively consistent responses. Notably, Probability Threshold method is exclusively employed in a double round form because we find some of the models struggle to generate both continuous probabilities and answers in a single round.

For the consistency based methods, we implemente a two-phase generation protocol: First, generating a reference answer with temperature = 0; Then sampling 10 variant answers with temperature = 1.0, with semantic equivalence between the basic answer and sampled answers evaluated by Qwen2.5-0.5B. With this process, we can get a consistency score between 0 to 10.

Specifically, for question rephrasing method, we leveraged Qwen2.5-7B to produce semantically equivalent question paraphrases. For the noised image method, we progressively added zero-mean Gaussian noise to the images during sampling, with the standard deviation incrementally increased from 0 in steps of 0.05. And for the cross model consistency method, we computed consistency scores using a combination of four responses generated by the primary model and three responses each from two other reference models.

\subsubsection{Complete Results}
Table~\ref{tab:qwencomplete}, Table~\ref{tab:llavacomplete} and Table~\ref{tab:deepseekcomplete} present the comprehensive performance evaluation of all methods across the three benchmark datasets and three LVLMs employed in our study.

\subsubsection{Observations and Analysis}
We proposed our mainly findings about LVLMs' knowledge boundary perception methods in Section~\ref{sec:LVLM performance}. Here, we discuss more detailed observations about them.

1) The Explain method improves alignment for both Deepseek-VL2 and Qwen2.5 when tested on the Dyn-VQA and Visual7W datasets. This demonstrates its effectiveness in enhancing LVLMs' knowledge boundary perception when processing relatively simple input questions.

2) The single-step Chain of Thought method effectively enhances alignment, whereas its double-step counterpart often leads to overconfidence and only marginally improves alignment for Qwen2.5-VL.

3) Both single-step and double-step Punish methods demonstrate limited effectiveness in mitigating overconfidence for Qwen2.5-VL and LLaVA-v1.5, as they fail to properly follow Punish Instructions.

4) Challenge method induces very high uncertain-rate in both three models, indicating that LVLMs are easily swayed by the output judgements.

5) For Qwen2.5-VL, rephrasing methods improve alignment on the Dyn-VQA dataset (language-focused), while the noise image method enhances performance on Visual7W (vision-focused). The combination of these two methods boosts alignment on the MMMU Pro dataset, which requires both language and vision comprehension. This reveals an interesting relationship between perturbation modalities and input query types.

\subsection{Comparing Perception between LVLMs and LLMs}
\label{app:compare}
While the main body presents a comparative analysis of single-step verbalization based confidence elicitation methods between LLMs and LVLMs, this section provides an extensive evaluation of: (i) double step verbalization based methods, (ii) answer consistency based methods, and (iii) token probability based method. The results can be found in Table~\ref{tab:LLM vs LVLM comprehensive}. The main observations are as follows.

\subsubsection{Double Step Verbalization Based Methods}
For double step verbalization based methods, the difference in performance between LLM and LVLM varies with the method.

1) For the Self-Judging method, Qwen2.5 exhibits higher alignment than Qwen2.5-VL. In contrast, the LLM counterparts of DeepSeek-VL2 and LLaVA tend to respond with ``certain'' to nearly all answers, resulting in extremely low consistency. This indicates a severe bias toward overconfident responses in these two LLMs.

2) For the Challenge method, LVLMs demonstrate higher uncertain-rates than LLMs, often approaching to near 1.0. This suggests that LVLMs are more likely to trust external judgments and consequently undermine their own decisions.

3) Under the Double-step Punish method, LLMs outperform LVLMs due to their stronger instruction following ability, achieving higher consistency and lower overconfidence.

\subsubsection{Answer Consistency Based Methods}

For answer consistency based methods, our observations are as follows:

1) Answer consistency based methods demonstrate superior alignment performance in LVLMs compared to LLMs.

2) DeepSeek-MoE exhibits strong consistency in its generated answers, maintaining high answer uniformity even when the outputs are incorrect. This behavior persists across both random sampling and rephrasing methods, leading to sustained overconfidence and suboptimal alignment performance.

3) The rephrasing strategy shows limited effectiveness in improving alignment metrics across all evaluated models, with the notable exception of Qwen2.5-VL. This observation holds true for both LVLMs and LLMs in our results.

\subsubsection{Token Probability Based Methods}
For the token probability based approach, as shown in Table~\ref{tab:LLM vs LVLM comprehensive}, our results reveal that LLMs exhibit relatively weaker confidence-accuracy alignment compared to LVLMs.



\subsection{Case Study of LVLM Outputs}

\subsubsection{CoT vs Img-CoT}

\begin{figure*}[ht!] 
\centering
\includegraphics[width=\linewidth]{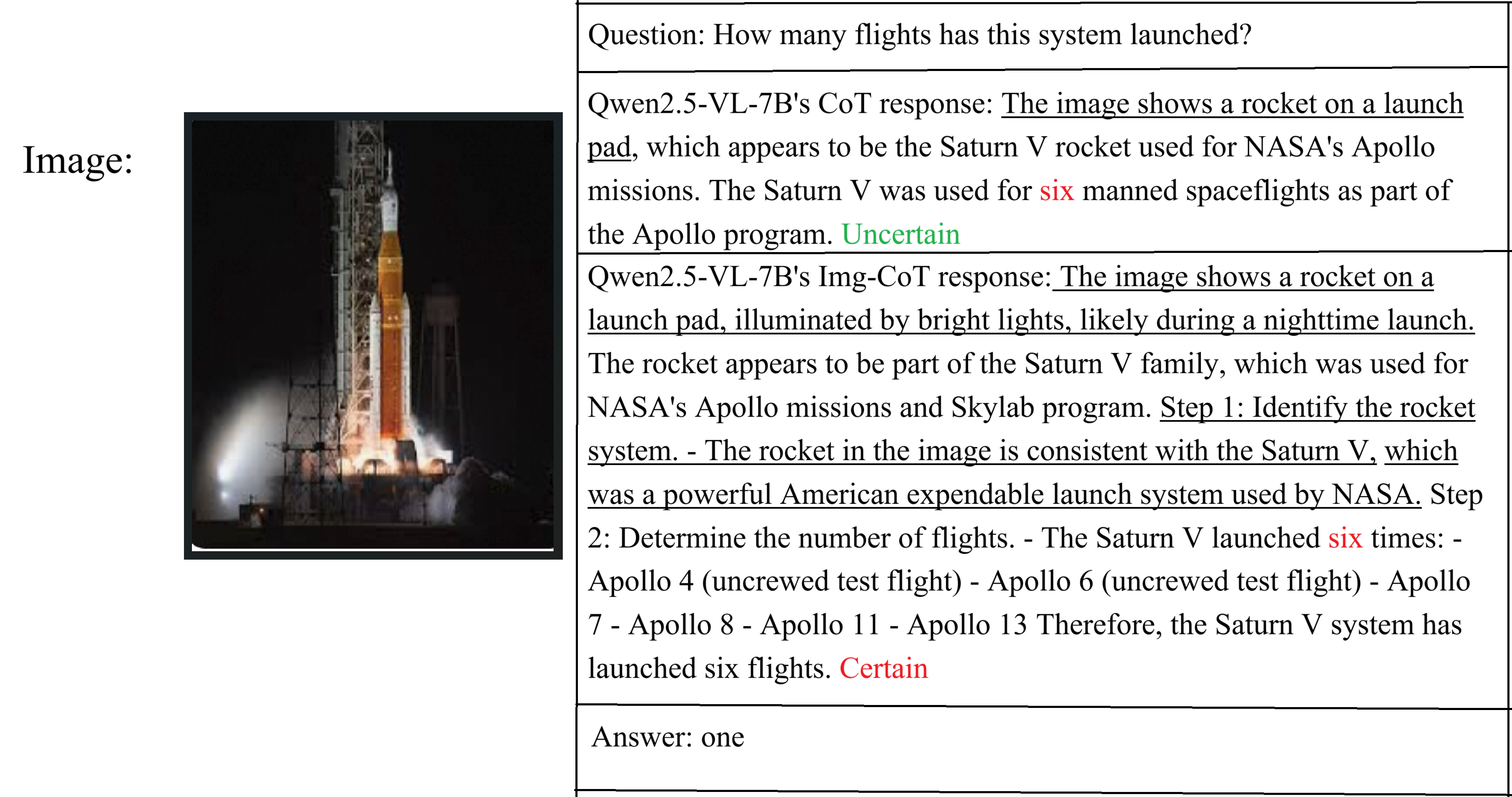}
\caption{Img-CoT and CoT methods' performance comparison. The description of the image is in \underline{underline}}
\label{fig:CoT vs Img-CoT}
\end{figure*}

As shown in Figure~\ref{fig:CoT vs Img-CoT}, In this case, the model generates extensive image descriptions under the Img-CoT method and confidently confirms its answer while the answer is wrong, demonstrating overconfidence.

\subsection{Small Scale LVLMs vs Large Scale LVLMs}
\begin{figure*}[ht!] 
\centering
\includegraphics[width=\linewidth]{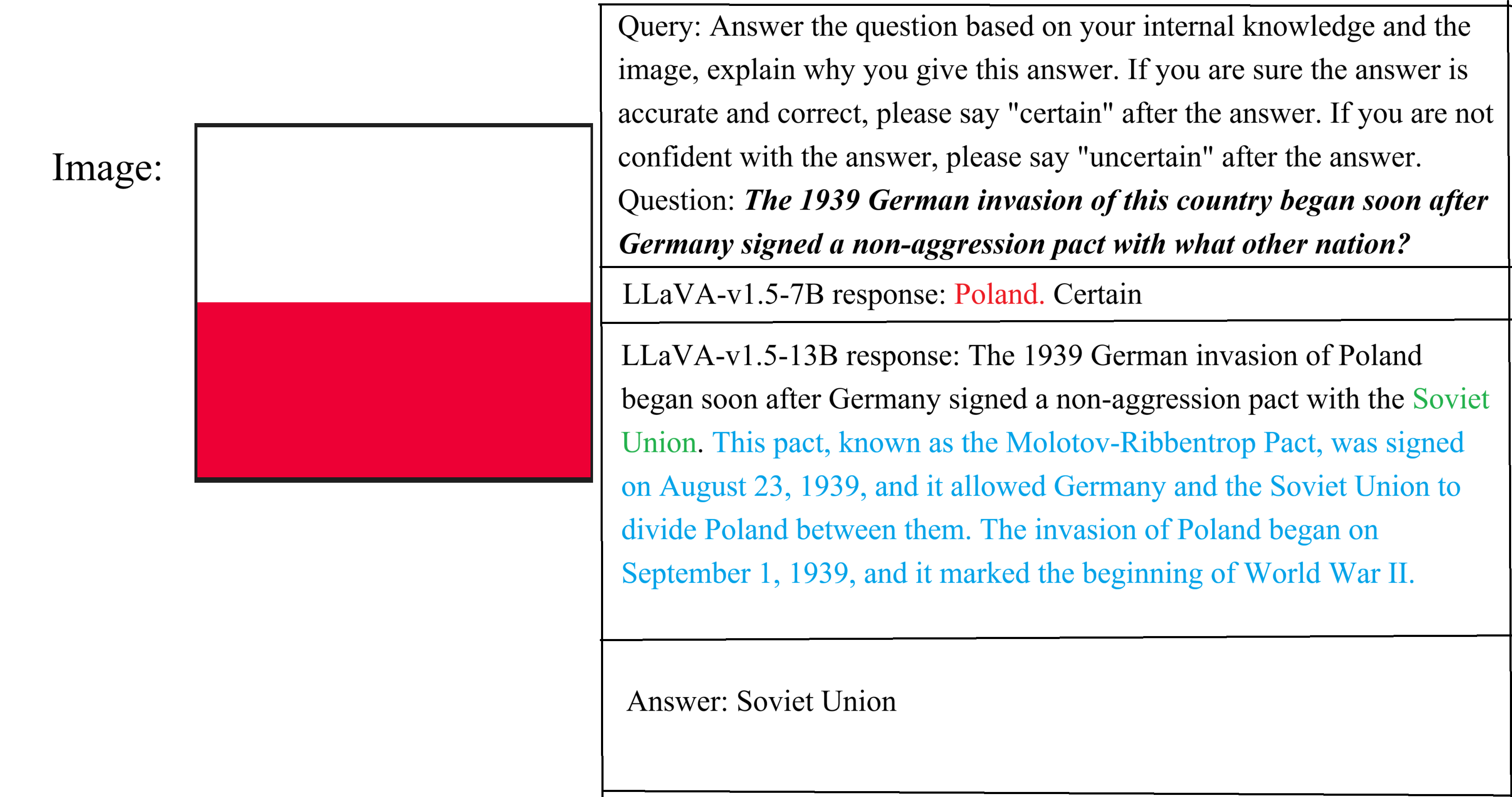}
\caption{Different scale LLaVA-v1.5's performance under the Explain instruction, The explanation of the answer is in blue.}
\label{fig:llava-instructfollowing}
\end{figure*}

As illustrated in Figure~\ref{fig:llava-instructfollowing}, we observe a notable difference in instruction following behavior between LLaVA-v1.5-7B and LLaVA-v1.5-14B. While the 7B model fails to adhere to the Explain instruction, the 14B variant successfully generates both explanatory sentences and the correct answer.

\begin{figure*}[ht!] 
\centering
\includegraphics[width=\linewidth]{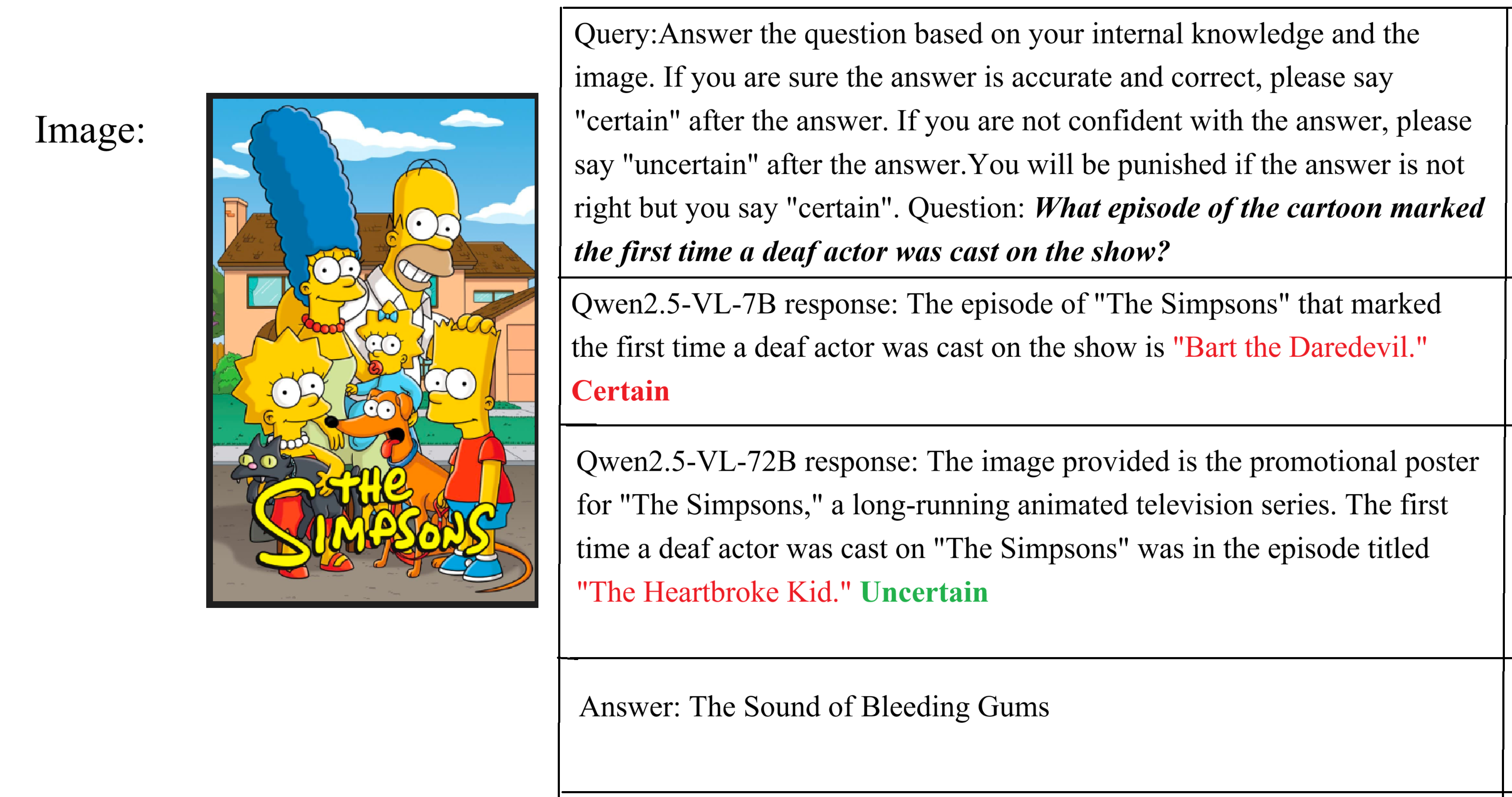}
\caption{Different scale Qwen2.5-VL's performance under the Punish instruction.}
\label{fig:qwen-instructfollowing}
\end{figure*}

Similarly, Figure~\ref{fig:qwen-instructfollowing} highlights a contrast in instruction following performance between Qwen2.5-7B and Qwen2.5-72B. The smaller 7B model exhibits overconfidence, incorrectly judging the answer without following the Punish instruction, whereas the 72B model responds more conservatively, aligning with the given directive.

\begin{table*}[ht]
\centering
\small
\caption{The performance of different methods on Qwen2.5-VL-7B-Instruct.}
\resizebox{\textwidth}{!}{
\begin{tabular}{@{}cccccc|ccccc|ccccc|@{}}
\toprule
\multirow{4}{*}{\textbf{method}} & \multicolumn{5}{c}{Dyn-VQA} & \multicolumn{5}{c}{Visual7W} & \multicolumn{5}{c}{MMMU Pro} \\
\cmidrule(lr){2-6} \cmidrule(lr){7-11} \cmidrule(l){12-16}
& \textbf{Unc-R.} & \textbf{Acc} & \textbf{Align.} & \textbf{Conser.} & \textbf{Overco.} & \textbf{Unc-R.} & \textbf{Acc} & \textbf{Align.} & \textbf{Conser.} & \textbf{Overco.} & \textbf{Unc-R.} & \textbf{Acc} & \textbf{Align.} & \textbf{Conser.} & \textbf{Overco.} \\
\midrule 

\colorbox{singlestep}{\makebox[\boxwidth]{\textbf{Vanilla}}} & \underline{0.7824} & 0.1846 & 0.7623 & \underline{0.1024} & 0.1353 & 0.3260 & 0.4380 & 0.5840 & 0.0900 & \underline{0.3260} & 0.3000 & 0.4564 & 0.4909 & \textbf{0.1327} & \underline{0.3764} \\
\colorbox{singlestep}{\makebox[\boxwidth]{\textbf{CoT}}} & 0.7276 & \textbf{0.2121} & \underline{0.7824} & 0.0786 & 0.1389 & \textbf{0.3840} & \underline{0.4920} & \underline{0.6080} & \textbf{0.1340} & 0.2580 & 0.2636 & \underline{0.6436} & \underline{0.6818} & 0.1127 & 0.2055 \\
\colorbox{singlestep}{\makebox[\boxwidth]{\textbf{Img-CoT}}} & 0.6545 & \underline{0.2048} & 0.7276 & 0.0658 & \underline{0.2066} & 0.2760 & \textbf{0.5020} & 0.6060 & 0.0860 & 0.3080 & \underline{0.2800} & \textbf{0.6636} & \textbf{0.7182} & 0.1127 & 0.1691 \\
\colorbox{singlestep}
{\makebox[\boxwidth]{\textbf{Punish}}} & 0.7112 & 0.1682 & 0.7112 & 0.0804 & \textbf{0.2084} & 0.2880 & 0.4280 & 0.5520 & 0.0820 & \textbf{0.3660} & 0.2691 & 0.4564 & 0.5000 & 0.1127 & \textbf{0.3873} \\
\colorbox{singlestep}{\makebox[\boxwidth]{\textbf{Explain}}} & \textbf{0.8282} & 0.1956 & \textbf{0.8117} & \textbf{0.1060} & 0.0823 & \underline{0.3640} & 0.4740 & \textbf{0.6180} & \underline{0.1100} & 0.2720 & \underline{0.2945} & 0.5309 & 0.5782 & \underline{0.1236} & 0.2982
 \\
\midrule

\colorbox{doublestep}{\makebox[\boxwidth]{\textbf{Self-Judging}}} & 0.1426 & 0.1883 & 0.3272 & 0.0018 & \textbf{0.6709} & 0.1100 & 0.4760 & 0.5500 & 0.0180 & \underline{0.4320} & 0.1882 & 0.5127 & \underline{0.5609} & 0.0701 & 0.3692
 \\
\colorbox{doublestep}{\makebox[\boxwidth]{\textbf{CoT}}} & \underline{0.5210} & 0.1883 & \underline{0.6435} & 0.0329 & 0.3236 & 0.1500 & 0.4760 & \underline{0.5700} & 0.0280 & 0.4020 & 0.2127 & 0.5127 & 0.5255 & 0.1000 & \underline{0.3745}
 \\
\colorbox{doublestep}{\makebox[\boxwidth]{\textbf{Challenge}}} & \textbf{0.9671} & 0.1883 & \textbf{0.8080} & \textbf{0.1737} & 0.0183 & \textbf{0.9800} & 0.4760 & 0.5280 & \textbf{0.4640} & 0.0080 & \textbf{0.9873} & 0.5127 & 0.4891 & \textbf{0.5055} & 0.0055
 \\
\colorbox{doublestep}{\makebox[\boxwidth]{\textbf{Punish}}} & 0.1426 & 0.1883 & 0.3272 & 0.0018 & \textbf{0.6709} & 0.0780 & 0.4760 & 0.5300 & 0.0120 & \textbf{0.4580} & 0.0436 & 0.5127 & 0.5164 & 0.0200 & \textbf{0.4636}
 \\
\colorbox{doublestep}{\makebox[\boxwidth]{\textbf{Prob-Thr}}} & 0.4991 & 0.1883 & 0.5960 & \underline{0.0457} & \underline{0.3583} & \underline{0.2140} & 0.4760 & \textbf{0.5820} & \underline{0.0540} & 0.3640 & \underline{0.4764} & 0.5127 & \textbf{0.5855} & \underline{0.2018} & 0.2127 \\
\midrule

\colorbox{consistency}{\makebox[\boxwidth]{\textbf{Random}}} & 0.4625 & 0.1883 & 0.5448 & 0.0530 & \textbf{0.4022} & \textbf{0.3020} & 0.4760 & 0.5700 & 0.1040 & \textbf{0.3260} & 0.4309 & 0.5127 & 0.5327 & 0.2055 & \underline{0.2618} \\
\colorbox{consistency}{\makebox[\boxwidth]{\textbf{Noised Img}}} & 0.7733 & 0.1883 & 0.7313 & 0.1152 & \underline{0.1536} & 0.4920 & 0.4760 & \underline{0.6000} & 0.1840 & 0.2160 & 0.3873 & 0.5127 & 0.5400 & 0.1800 & \textbf{0.2800} \\
\colorbox{consistency}{\makebox[\boxwidth]{\textbf{Rephr}}} & \textbf{0.9543} & 0.1883 & \underline{0.8026} & \textbf{0.1700} & 0.0274 & 0.4340 & 0.4760 & 0.5660 & 0.1720 & \underline{0.2620} & \textbf{0.5655} & 0.5127 & 0.5364 & \textbf{0.2709} & 0.1927 \\
\colorbox{consistency}{\makebox[\boxwidth]{\textbf{Reph+Nois}}} & 0.8958 & 0.1883 & 0.7733 & 0.1554 & 0.0713 & \underline{0.4940} & 0.4760 & 0.5500 & \textbf{0.2100} & 0.2400 & 0.4418 & 0.5127 & \underline{0.5509} & 0.2018 & 0.2473 \\
\colorbox{consistency}{\makebox[\boxwidth]{\textbf{Cross Model}}} & \underline{0.9469} & 0.1883 & \textbf{0.8208} & \underline{0.1572} & 0.0219 & \textbf{0.5720} & 0.4760 & \textbf{0.6320} & \underline{0.2080} & 0.1600 & \underline{0.5036} & 0.5127 & \textbf{0.5800} & \underline{0.2182} & 0.2018 \\
\midrule
\colorbox{PPL}{\makebox[\boxwidth]{\textbf{PPL Thr}}} & 0.8885 & 0.1993 & 0.7916 & 0.1481 & 0.0603 & 0.8060 & 0.4760 & 0.6020 & 0.3400 & 0.0580 & 0.9436 & 0.4091 & 0.6073 & 0.3727 & 0.0200 \\

\bottomrule
\end{tabular}}

\label{tab:qwencomplete}
\end{table*}

\begin{table*}[ht!]
\centering
\small
\caption{The performance of different methods on LLaVA-v1.5-7B.}
\resizebox{\textwidth}{!}{
\begin{tabular}{@{}cccccc|ccccc|ccccc|@{}}
\toprule
\multirow{4}{*}{\textbf{method}} & \multicolumn{5}{c}{Dyn-VQA} & \multicolumn{5}{c}{Visual7W} & \multicolumn{5}{c}{MMMU Pro} \\
\cmidrule(lr){2-6} \cmidrule(lr){7-11} \cmidrule(l){12-16}
& \textbf{Unc-R.} & \textbf{Acc} & \textbf{Align.} & \textbf{Conser.} & \textbf{Overco.} & \textbf{Unc-R.} & \textbf{Acc} & \textbf{Align.} & \textbf{Conser.} & \textbf{Overco.} & \textbf{Unc-R.} & \textbf{Acc} & \textbf{Align.} & \textbf{Conser.} & \textbf{Overco.} \\
\midrule

\colorbox{singlestep}{\makebox[\boxwidth]{\textbf{Vanilla}}} & 0.4899 & 0.0878 & 0.5338 & 0.0219 & 0.4442 & \textbf{0.0260} & 0.3920 & 0.4140 & 0.0020 & 0.5840 & 0.0855 & 0.2018 & 0.2509 & 0.0182 & 0.7309 \\
\colorbox{singlestep}{\makebox[\boxwidth]{\textbf{CoT}}} & \underline{0.5119} & 0.0841 & \underline{0.5375} & \underline{0.0311} & 0.4314 & 0.0220 & 0.3840 & 0.3940 & \textbf{0.0060} & \underline{0.6000} & 0.1418 & 0.1545 & 0.2418 & 0.0273 & \textbf{0.7309} \\
\colorbox{singlestep}{\makebox[\boxwidth]{\textbf{Img-CoT}}} & \textbf{0.5265} & \underline{0.0914} & \textbf{0.5484} & \textbf{0.0347} & 0.4168 & 0.0180 & \textbf{0.4000} & \underline{0.4140} & 0.0020 & 0.5840 & \underline{0.1527} & \underline{0.2527} & \underline{0.2964} & \underline{0.0545} & 0.6491 \\
\colorbox{singlestep}
{\makebox[\boxwidth]{\textbf{Punish}}} & 0.4497 & \textbf{0.0951} & 0.4899 & 0.0274 & \underline{0.4826} & \underline{0.0260} & \underline{0.3960} & \textbf{0.4180} & 0.0020 & 0.5800 & \textbf{0.2291} & \textbf{0.2727} & \textbf{0.3745} & \textbf{0.0636} & 0.5618 \\
\colorbox{singlestep}{\makebox[\boxwidth]{\textbf{Explain}}} & 0.4205 & 0.0841 & 0.4534 & 0.0274 & \textbf{0.5192} & 0.0100 & 0.3840 & 0.3900 & 0.0020 & \textbf{0.6080} & 0.0727 & 0.1709 & 0.2109 & 0.0164 & \underline{0.7727} \\
\midrule

\colorbox{doublestep}{\makebox[\boxwidth]{\textbf{Self-Judging}}} & \underline{0.1718} & 0.1005 & \underline{0.2468} & \underline{0.0128} & 0.7404 & \underline{0.0020} & 0.4200 & \underline{0.4220} & 0.0000 & \underline{0.5780} & \underline{0.0109} & 0.3218 & \underline{0.3327} & 0.0000 & \underline{0.6673} \\
\colorbox{doublestep}{\makebox[\boxwidth]{\textbf{CoT}}} & 0.0494 & 0.1005 & 0.1463 & 0.0018 & \underline{0.8519} & 0.0000 & 0.4200 & 0.4200 & 0.0000 & \textbf{0.5800} & 0.0000 & 0.3218 & 0.3218 & 0.0000 & \textbf{0.6782} \\
\colorbox{doublestep}{\makebox[\boxwidth]{\textbf{Challenge}}} & \sout{1.0000} & \sout{0.1005} & \sout{0.8995} & \sout{0.1005} & \sout{0.0000} & \sout{1.0000} & \sout{0.4200} & \sout{0.5800} & \sout{0.4200} & \sout{0.0000} & \sout{1.0000} & \sout{0.3218} & \sout{0.6782} & \sout{0.3218} & \sout{0.0000} \\ 
\colorbox{doublestep}{\makebox[\boxwidth]{\textbf{Punish}}} & 0.0293 & 0.1005 & 0.1298 & 0.0000 & \textbf{0.8702} & 0.0000 & 0.4200 & 0.4200 & 0.0000 & \textbf{0.5800} & 0.0000 & 0.3218 & 0.3218 & 0.0000 & \textbf{0.6782} \\
\colorbox{doublestep}{\makebox[\boxwidth]{\textbf{Prob-Thr}}} & \textbf{0.8464} & 0.1005 & \textbf{0.7971} & \textbf{0.0750} & 0.1280 & \textbf{0.6780} & 0.4200 & \textbf{0.6140} & \textbf{0.2420} & 0.1440 & \textbf{0.7964} & 0.3218 & \textbf{0.6091} & \textbf{0.2545} & 0.1364 \\
\midrule

\colorbox{consistency}{\makebox[\boxwidth]{\textbf{Random}}} & 0.9872 & 0.1005 & \underline{0.8976} & 0.0951 & \textbf{0.0073} & \underline{0.6680} & 0.4200 & \textbf{0.7080} & 0.1900 & 0.1020 & \underline{0.9745} & 0.3218 & \textbf{0.6709} & 0.3127 & 0.0164 \\
\colorbox{consistency}{\makebox[\boxwidth]{\textbf{Noised Img}}} & 0.9963 & 0.1005 & 0.8958 & \textbf{0.1005} & \underline{0.0037} & 0.5660 & 0.4200 & 0.6740 & 0.1560 & 0.1700 & \textbf{0.9836} & 0.3218 & 0.6655 & \textbf{0.3200} & 0.0145 \\
\colorbox{consistency}{\makebox[\boxwidth]{\textbf{Rephr}}} & \underline{0.9981} & 0.1005 & \underline{0.8976} & \textbf{0.1005} & 0.0018 & 0.6560 & 0.4200 & \underline{0.6920} & \underline{0.1920} & \underline{0.1160} & 0.9345 & 0.3218 & \underline{0.6672} & 0.2945 & \textbf{0.0382} \\
\colorbox{consistency}{\makebox[\boxwidth]{\textbf{Reph+Nois}}} & \textbf{0.9982} & 0.1005 & \textbf{0.9013} & \underline{0.0987} & 0.0000 & \textbf{0.7020} & 0.4200 & 0.6780 & \textbf{0.2220} & 0.1000 & 0.9655 & 0.3218 & 0.6655 & 0.3109 & \underline{0.0236} \\
\colorbox{consistency}{\makebox[\boxwidth]{\textbf{Cross Model}}} & \textbf{0.9982} & 0.1005 & \underline{0.8976} & \textbf{0.1005} & 0.0018 & 0.5320 & 0.4200 & 0.6520 & 0.1500 & \textbf{0.1980} & 0.9727 & 0.3218 & 0.6618 & \underline{0.3164} & 0.0218 \\

\midrule
\colorbox{PPL}{\makebox[\boxwidth]{\textbf{PPL Thr}}} & 0.8903 & 0.1005 & \underline{0.8519} & \textbf{0.0695} & 0.0786 & 0.5860 & 0.4200 & 0.7060 & 0.1460 & 0.1380 & 0.9727 & 0.3218 & 0.6800 & 0.3073 & 0.0127 \\

\bottomrule
\end{tabular}}

\label{tab:llavacomplete}
\end{table*}

\begin{table*}[ht!]
\centering
\small
\caption{The performance of different methods on DeepSeek-VL2-16B.}
\resizebox{\textwidth}{!}{
\begin{tabular}{@{}cccccc|ccccc|ccccc|@{}}
\toprule
\multirow{4}{*}{\textbf{method}} & \multicolumn{5}{c}{Dyn-VQA} & \multicolumn{5}{c}{Visual7W} & \multicolumn{5}{c}{MMMU Pro} \\
\cmidrule(lr){2-6} \cmidrule(lr){7-11} \cmidrule(l){12-16}
& \textbf{Unc-R.} & \textbf{Acc} & \textbf{Align.} & \textbf{Conser.} & \textbf{Overco.} & \textbf{Unc-R.} & \textbf{Acc} & \textbf{Align.} & \textbf{Conser.} & \textbf{Overco.} & \textbf{Unc-R.} & \textbf{Acc} & \textbf{Align.} & \textbf{Conser.} & \textbf{Overco.} \\
\midrule

\colorbox{singlestep}{\makebox[\boxwidth]{\textbf{Vanilla}}} & \underline{0.7879} & 0.1463 & 0.6527 & \underline{0.1408} & 0.2066 & \underline{0.4120} & 0.1840 & 0.2820 & \underline{0.1580} & \textbf{0.5600} & \underline{0.4091} & 0.2673 & 0.2727 & \underline{0.2018} & 0.5255 \\
\colorbox{singlestep}{\makebox[\boxwidth]{\textbf{CoT}}} & 0.6380 & \underline{0.1700} & 0.6362 & 0.0859 & \underline{0.2779} & 0.1780 & 0.4600 & \underline{0.5540} & 0.0420 & 0.4040 & 0.0873 & \underline{0.3509} & \underline{0.3836} & 0.0273 & \textbf{0.5891} \\
\colorbox{singlestep}{\makebox[\boxwidth]{\textbf{Img-CoT}}} & 0.5356 & \textbf{0.2011} & 0.6344 & 0.0512 & \textbf{0.3144} & 0.0640 & \textbf{0.4960} & 0.5360 & 0.0120 & 0.4520 & 0.1055 & \textbf{0.4582} & \textbf{0.5236} & 0.0200 & 0.4564 \\
\colorbox{singlestep}
{\makebox[\boxwidth]{\textbf{Punish}}} & \textbf{0.8483} & 0.1609 & \textbf{0.7093} & \textbf{0.1499} & 0.1407 & \textbf{0.4580} & 0.2680 & 0.3500 & \textbf{0.1880} & \underline{0.4620} & \textbf{0.4782} & 0.3054 & 0.3145 & \textbf{0.2345} & 0.4509 \\
\colorbox{singlestep}{\makebox[\boxwidth]{\textbf{Explain}}} & 0.7861 & 0.1682 & \underline{0.6984} & 0.1280 & 0.1737 & 0.2780 & \underline{0.4640} & \textbf{0.5700} & 0.0860 & 0.3440 & 0.2073 & 0.3382 & 0.3491 & 0.0982 & \underline{0.5527} \\
\midrule

\colorbox{doublestep}{\makebox[\boxwidth]{\textbf{Self-Judging}}} & 0.0018 & 0.1974 & 0.1993 & 0.0000 & \textbf{0.8007} & 0.0020 & 0.4760 & 0.4780 & 0.0000 & \underline{0.5220} & 0.0018 & 0.4255 & 0.4236 & 0.0018 & \textbf{0.5745} \\
\colorbox{doublestep}{\makebox[\boxwidth]{\textbf{CoT}}} & 0.0055 & 0.1974 & 0.2029 & 0.0000 & \underline{0.7971} & 0.0000 & 0.4760 & 0.4760 & 0.0000 & \textbf{0.5240} & 0.0073 & 0.4255 & 0.4255 & 0.0036 & \underline{0.5709} \\
\colorbox{doublestep}{\makebox[\boxwidth]{\textbf{Challenge}}} & \textbf{0.9945} & 0.1974 & \textbf{0.8007} & \textbf{0.1956} & 0.0037 & \textbf{0.9960} & 0.4760 & 0.5240 & \textbf{0.4740} & 0.0020 & \textbf{0.9309} & 0.4255 & \textbf{0.5709} & \textbf{0.3927} & 0.0364 \\
\colorbox{doublestep}{\makebox[\boxwidth]{\textbf{Punish}}} & 0.3144 & 0.1974 & 0.4936 & 0.0091 & 0.4973 & 0.0620 & 0.4760 & \underline{0.5300} & 0.0040 & 0.4660 & 0.0273 & 0.4255 & 0.4345 & 0.0091 & 0.5564 \\ 
\colorbox{doublestep}{\makebox[\boxwidth]{\textbf{Prob-Thr}}} & \underline{0.7239} & 0.1974 & \underline{0.6910} & \underline{0.1152} & 0.1938 & \underline{0.7280} & 0.4760 & \textbf{0.6060} & \underline{0.2980} & 0.0960 & \underline{0.7473} & 0.4255 & \underline{0.5218} & \underline{0.3127} & 0.1655 \\
\midrule

\colorbox{consistency}{\makebox[\boxwidth]{\textbf{Random}}} & \textbf{0.9963} & 0.1974 & 0.8026 & \textbf{0.1956} & \underline{0.0018} & \underline{0.5800} & 0.4740 & \underline{0.6460} & \textbf{0.2040} & 0.1500 & \textbf{0.8509} & 0.4180 & 0.6000 & \textbf{0.3345} & 0.0655 \\
\colorbox{consistency}{\makebox[\boxwidth]{\textbf{Noised Img}}} & \underline{0.9927} & 0.1974 & 0.8062 & 0.1920 & \underline{0.0018} & 0.5480 & 0.4740 & 0.6300 & \underline{0.1960} & 0.1740 & 0.7418 & 0.4180 & \underline{0.5818} & 0.2891 & \textbf{0.1291} \\
\colorbox{consistency}{\makebox[\boxwidth]{\textbf{Rephr}}} & 0.9689 & 0.1974 & \underline{0.8080} & 0.1792 & 0.0127 & 0.4480 & 0.4740 & 0.6260 & 0.1480 & \textbf{0.2260} & 0.7982 & 0.4180 & 0.5764 & 0.3200 & 0.1036 \\
\colorbox{consistency}{\makebox[\boxwidth]{\textbf{Reph+Nois}}} & 0.9670 & 0.1974 & \textbf{0.8099} & 0.1773 & \textbf{0.0128} & 0.5140 & 0.4740 & 0.6120 & 0.1880 & \underline{0.2000} & 0.7945 & 0.4180 & 0.5618 & 0.3255 & \underline{0.1127} \\
\colorbox{consistency}{\makebox[\boxwidth]{\textbf{Cross Model}}} & \textbf{0.9963} & 0.1974 & 0.8062 & \underline{0.1938} & 0.0000 & \textbf{0.6080} & 0.4740 & \textbf{0.6740} & \textbf{0.2040} & 0.1220 & \underline{0.8327} & 0.4180 & \textbf{0.5964} & \underline{0.3273} & 0.0764 \\

\midrule
\colorbox{PPL}{\makebox[\boxwidth]{\textbf{PPL Thr}}} & 0.8958 & 0.1974 & 0.7934 & 0.1499 & 0.0567 & 0.5500 & 0.4780 & 0.6280 & 0.2000 & 0.1720 & 0.9418 & 0.4436 & 0.5345 & 0.4255 & 0.0400 \\

\bottomrule
\end{tabular}}

\label{tab:deepseekcomplete}
\end{table*}

\begin{table*}[t]
\centering
\small
\caption{Performance comparison of double step verbaliztion based methods, consistency based methods and answer consistency based methods on the Dyn-VQA dataset: LVLMs vs. LLMs}
\resizebox{\textwidth}{!}{
\begin{tabular}{@{}ccccccc|ccccc|ccccc|@{}}
\toprule

\multirow{4}{*}{\textbf{method}} & \multirow{4}{*}{\textbf{Model Type}} & \multicolumn{5}{c}{Qwen2.5} & \multicolumn{5}{c}{LLaVA1.5} & \multicolumn{5}{c}{DeepSeek-VL2}\\
\cmidrule(lr){3-7} \cmidrule(lr){8-12} \cmidrule(l){13-17}
& &  \textbf{Unc-R.} & \textbf{Acc} & \textbf{Align.} & \textbf{Conser.} & \textbf{Overco.} & \textbf{Unc-R.} & \textbf{Acc} & \textbf{Align.} & \textbf{Conser.} & \textbf{Overco.} & \textbf{Unc-R.} & \textbf{Acc} & \textbf{Align.} & \textbf{Conser.} & \textbf{Overco.} \\
\midrule
\multirow{2}{*}{\colorbox{doublestep}{\makebox[\boxwidth]{\textbf{Self-Judging}}}} & LVLM & 0.1426 & 0.1883 & 0.3272 & 0.0018 & \textbf{0.6709} & \textbf{0.0018} & 0.1974 & 0.1993 & 0.0000 & \textbf{0.8007} & \textbf{0.1718} & 0.1005 & \textbf{0.2468} & \textbf{0.0128} & 0.7404 \\
& LLM & \textbf{0.2943} & \textbf{0.2998} & \textbf{0.5649} & \textbf{0.0146} & 0.4205 & 0.0000 & \textbf{0.2962} & \textbf{0.2962} & 0.0000 & 0.7038 & 0.0000 & \textbf{0.2139} & 0.2139 & 0.0000 & \textbf{0.7861} \\
\midrule
\multirow{2}{*}{\colorbox{doublestep}{\makebox[\boxwidth]{\textbf{CoT}}}} & LVLM & \textbf{0.5210} & 0.1883 & \textbf{0.6435} & \textbf{0.0329} & 0.3236 & 0.0055 & 0.1974 & 0.2029 & 0.0000 & \textbf{0.7971} & 0.0494 & 0.1005 & 0.1463 & 0.0018 & \textbf{0.8519} \\
& LLM & 0.2925 & \textbf{0.2998} & 0.5411 & 0.0256 & \textbf{0.4333} & \textbf{0.2888} & \textbf{0.2962} & \textbf{0.5192} & \textbf{0.0329} & 0.4479 & \textbf{0.2761} & \textbf{0.2139} & \textbf{0.4680} & \textbf{0.0110} & 0.5210 \\
\midrule
\multirow{2}{*}{\colorbox{doublestep}{\makebox[\boxwidth]{\textbf{Challenge}}}} & LVLM & \textbf{0.9671} & 0.1883 & \textbf{0.8080} & \textbf{0.1737} & 0.0183 & \textbf{0.9945} & 0.1974 & \textbf{0.8007} & 0.1956 & 0.0037 & \textbf{1.0000} & 0.1005 & \textbf{0.8995} & 0.1005 & 0.0000 \\
& LLM & 0.7148 & \textbf{0.2998} & 0.7514 & 0.1316 & \textbf{0.1170} & 0.8684 & \textbf{0.2962} & 0.6563 & \textbf{0.2541} & \textbf{0.0896} & 0.9853 & \textbf{0.2139} & 0.7898 & \textbf{0.2048} & \textbf{0.0055} \\
\midrule
\multirow{2}{*}{\colorbox{doublestep}{\makebox[\boxwidth]{\textbf{Punish}}}} & LVLM & 0.1426 & 0.1883 & 0.3272 & 0.0018 & \textbf{0.6709} & \textbf{0.3144} & 0.1974 & 0.4936 & 0.0091 & \textbf{0.4973} & 0.0293 & 0.1005 & 0.1298 & 0.0000 & \textbf{0.8702} \\ 
& LLM & \textbf{0.5448} & \textbf{0.2998} & \textbf{0.6910} & \textbf{0.0768} & 0.2322 & 0.2852 & \textbf{0.2962} & \textbf{0.5302} & \textbf{0.0256} & 0.4442 & \textbf{0.1974} & \textbf{0.2139} & \textbf{0.4113} & 0.0000 & 0.5887 \\
\midrule
\multirow{2}{*}{\colorbox{doublestep}{\makebox[\boxwidth]{\textbf{Prob-Thr}}}} & LVLM & 0.4991 & 0.1883 & 0.5960 & 0.0457 & \textbf{0.3583} & \textbf{0.7239} & 0.1974 & \textbf{0.6910} & \textbf{0.1152} & 0.1938 & 0.8464 & 0.1005 & \textbf{0.7971} & 0.0750 & \textbf{0.1280} \\
& LLM & \textbf{0.5941} & \textbf{0.2998} & \textbf{0.6709} & \textbf{0.1115} & 0.2175 & 0.1773 & \textbf{0.2962} & 0.4333 & 0.0201 & \textbf{0.5466} & \textbf{0.9963} & \textbf{0.2139} & 0.7824 & \textbf{0.2139} & 0.0037 \\
\midrule
\multirow{2}{*}{\colorbox{consistency}{\makebox[\boxwidth]{\textbf{Random}}}} & LVLM & 0.4625 & 0.1883 & 0.5448 & 0.0530 & \textbf{0.4022} & \textbf{0.9963} & 0.1974 & \textbf{0.8026} & \textbf{0.1956} & 0.0018 & \textbf{0.9872} & 0.1005 & \textbf{0.8976} & 0.0951 & 0.0073 \\ 
& LLM & \textbf{0.9287} & \textbf{0.2998} & \textbf{0.7203} & \textbf{0.2541} & 0.0256 & 0.4863 & \textbf{0.2962} & 0.5448 & 0.1188 & \textbf{0.3364} & 0.8921 & \textbf{0.2139} & 0.7806 & \textbf{0.1627} & \textbf{0.0567} \\
\midrule
\multirow{2}{*}{\colorbox{consistency}{\makebox[\boxwidth]{\textbf{Rephr}}}} & LVLM & \textbf{0.9543} & 0.1883 & \textbf{0.8026} & 0.1700 & 0.0274 & \textbf{0.9689} & 0.1974 & \textbf{0.8080} & \textbf{0.1792} & 0.0127 & \textbf{0.9981} & 0.1005 & \textbf{0.8976} & 0.1005 & 0.0018 \\
& LLM & 0.9068 & \textbf{0.2998} & 0.7203 & \textbf{0.2431} & \textbf{0.0366} & 0.4991 & \textbf{0.2962} & 0.5539 & 0.1207 & \textbf{0.3254} & 0.8757 & \textbf{0.2139} & 0.7751 & \textbf{0.1572} & \textbf{0.0676} \\
\midrule
\multirow{2}{*}{\colorbox{PPL}{\makebox[\boxwidth]{\textbf{PPL Thr}}}} & LVLM & \textbf{0.8885} & 0.1993 & \textbf{0.7916} & 0.1481 & \textbf{0.0603} & \textbf{0.8903} & 0.1005 & \textbf{0.8519} & 0.0695 & 0.0786 & \textbf{0.8958} & 0.1974 & \textbf{0.7934} & \textbf{0.1499} & 0.0567 \\
& LLM & 0.8519 & \textbf{0.3217} & 0.7313 & \textbf{0.2212} & 0.0475 & 0.7587 & \textbf{0.2980} & 0.6837 & \textbf{0.1865} & \textbf{0.1298} & 0.7458 & \textbf{0.2121} & 0.7422 & 0.1079 & \textbf{0.1499} \\
\bottomrule
\end{tabular}}

\label{tab:LLM vs LVLM comprehensive}
\end{table*}

\end{document}